\documentclass[journal]{IEEEtran}

\PassOptionsToPackage{hyphens}{url}
\PassOptionsToPackage{bookmarks=false}{hyperref}

\usepackage{orcidlink}

\newcommand{\IEEEauthororcidlink}[1]{\orcidlink{#1}}

\usepackage[utf8]{inputenc}
\usepackage[T1]{fontenc}

\usepackage{xurl}

\usepackage{microtype}

\usepackage{dblfloatfix}

\usepackage{amsmath,amssymb,mathtools}
\usepackage{bm}
\usepackage{booktabs}
\usepackage{multirow}
\usepackage{array}
\usepackage{siunitx}
\usepackage{threeparttable}
\usepackage{threeparttablex}
\usepackage{tabularx}
\usepackage{makecell}
\usepackage{pbox}
\usepackage{algorithm}
\usepackage{algorithmic}

\newcolumntype{L}[1]{>{\raggedright\arraybackslash}p{#1}}
\newcolumntype{C}[1]{>{\centering\arraybackslash}p{#1}}
\newcolumntype{R}[1]{>{\raggedleft\arraybackslash}p{#1}}
\newcolumntype{Y}{>{\raggedright\arraybackslash}X}

\PassOptionsToPackage{final}{graphicx}   
\usepackage{graphicx}
\setkeys{Gin}{draft=false}

\graphicspath{{./}}
\DeclareGraphicsExtensions{.pdf,.png,.jpg,.jpeg}
\usepackage[caption=false,font=footnotesize]{subfig}

\usepackage[dvipsnames,table]{xcolor}

\usepackage{placeins}
\usepackage{balance}

\usepackage{cite}
\usepackage{url}

\usepackage{hyperref}

\hypersetup{
    colorlinks=true,
    linkcolor=black,
    citecolor=NavyBlue,
    urlcolor=black
}

\usepackage[capitalise,noabbrev,nameinlink]{cleveref}

\DeclareSIUnit\rpm{rpm}
\DeclareSIUnit\db{dB}
\DeclareSIUnit\fps{fps}
\DeclareSIUnit\ms{ms}
\begin{document}

\title{PI-TTA: Physics-Informed Source-Free Test-Time Adaptation for Robust Human Activity Recognition on Mobile Devices}
\author{%
  Changyu~Li$^{1}$\IEEEauthororcidlink{0009-0007-0876-0310},
  Lu~Wang$^{5}$\IEEEauthororcidlink{0000-0001-6345-3873},~\IEEEmembership{Senior Member,~IEEE},
  Ming~Lei$^{1}$\IEEEauthororcidlink{0009-0007-3989-6755},
  Jiashen~Liu$^{2}$\IEEEauthororcidlink{0009-0000-0908-0877},
  Yichen~Zhang$^{4}$,
  Kaishun~Wu$^{3}$\IEEEauthororcidlink{0000-0003-2216-0737},~\IEEEmembership{Fellow,~IEEE},
  and Fei~Luo$^{1}$\IEEEauthororcidlink{0000-0001-9760-1520}%
\thanks{$^{1}$Changyu~Li, Ming~Lei, and Fei~Luo are with the School of Computing and Information Technology, Great Bay University, Dongguan, China (e-mail: Changyuli.021230@gmail.com; leiming61@hrbeu.edu.cn; luofei2018@outlook.com).}%
\thanks{$^{2}$Jiashen~Liu is with the University of Warwick, Coventry, U.K. (e-mail: Jiashen.Liu@warwick.ac.uk).}%
\thanks{$^{3}$Kaishun~Wu is with the Hong Kong University of Science and Technology (Guangzhou), Guangzhou, China (e-mail: wuks@hkust-gz.edu.cn).}%
\thanks{$^{4}$Yichen~Zhang is with The University of Queensland, Brisbane, Australia (e-mail: yczhang0713@126.com).}
\thanks{$^{5}$Lu~Wang is with the College of Computer Science and Software Engineering, Shenzhen University, Shenzhen, China (e-mail: wanglu@szu.edu.cn).}%
}
\maketitle
\begin{abstract}
Source-free test-time adaptation (TTA) is appealing for mobile and wearable sensing because it enables on-device personalization from unlabeled test streams without centralizing private data. However, sensor-based human activity recognition (HAR) poses challenges that are less pronounced in standard vision benchmarks: behavioral inertial streams are temporally correlated and often exhibit within-session shifts caused by sensor rotation, placement change, and sampling-rate drift. Under this streaming non-i.i.d.\ setting, widely used vision-style TTA objectives can become unstable, leading to overconfident errors, representation collapse, and catastrophic forgetting. We propose \textsc{PI-TTA}, a lightweight source-free adaptation framework that stabilizes online updates through three physics-consistent constraints: gravity consistency, short-horizon temporal continuity, and spectral stability. \textsc{PI-TTA} updates the same small parameter subset as strong source-free baselines and incurs only modest overhead, making it suitable for on-device deployment. Experiments on USCHAD, PAMAP2, and mHealth under long-sequence stress tests and factorized shift protocols show that \textsc{PI-TTA} mitigates the severe degradation observed in confidence-driven baselines and preserves stable adaptation under sustained streaming conditions. It improves long-sequence accuracy by up to 9.13\% and reduces physical-violation rates by 27.5\%, 24.1\%, and 45.4\% on USCHAD, PAMAP2, and mHealth, respectively. These results demonstrate that physics-informed adaptation can improve accuracy, stability, and deployment reliability for real-world mobile sensing systems.
\end{abstract}

\section{Introduction}
\IEEEPARstart{M}{obile} and wearable sensing systems have become an important infrastructure for behavioral understanding applications such as human activity recognition (HAR), digital health monitoring, and context-aware human--computer interaction~\cite{jiang2015wearablecnn,tang2022tcdattention}. Although recent deep models and cross-domain learning methods have substantially improved HAR accuracy, reliable deployment on mobile devices remains challenging. In practice, the challenge is not only to recognize activities under distribution shift, but to do so under strict on-device latency, memory, energy, and sustained runtime constraints~\cite{chang2020uda_har,qin2019asttl,chen2020metier,bai2020damun,shi2016edge,lim2020fedcomst,alfarra2024ttatime,luo2025bideepvit,luo2025activitymamba}. This creates a fundamental tension: the model should adapt to evolving test conditions, yet the adaptation process itself must remain lightweight and stable for continuous mobile execution.

This tension is particularly pronounced for behavioral inertial streams. During a single deployment session, sensors may rotate inside pockets, device placement may shift across body locations such as the waist, arm, or chest, and acquisition characteristics may drift over time~\cite{sztyler2016onbody,chang2020uda_har,qin2019asttl,chen2020metier,bai2020damun,kwon2020imutube}. As a result, the deployment stream is temporally correlated and non-i.i.d., rather than a collection of independent test samples. This differs fundamentally from shuffled offline evaluation. In realistic mobile use, the model is repeatedly queried and updated under evolving stream conditions, and any adaptation error can propagate across subsequent windows rather than being averaged out over independent examples.

Source-free test-time adaptation (TTA) is therefore an appealing paradigm for mobile HAR. It enables the model to adapt directly on unlabeled deployment streams without storing or centralizing private source data, which is attractive for privacy-sensitive wearable applications. However, this setting imposes requirements that are stricter than those of conventional offline robustness. The adaptation mechanism must remain lightweight, online, and resilient to long correlated segments, because recovery from a poor update is costly under limited compute and energy budgets~\cite{alfarra2024ttatime,shi2016edge,luo2025bideepvit,luo2025activitymamba}. To our knowledge, deployment-constrained source-free adaptation on temporally correlated sensor streams remains underexplored in HAR.

\begin{figure*}[t]
\centering
\includegraphics[width=0.85\textwidth]{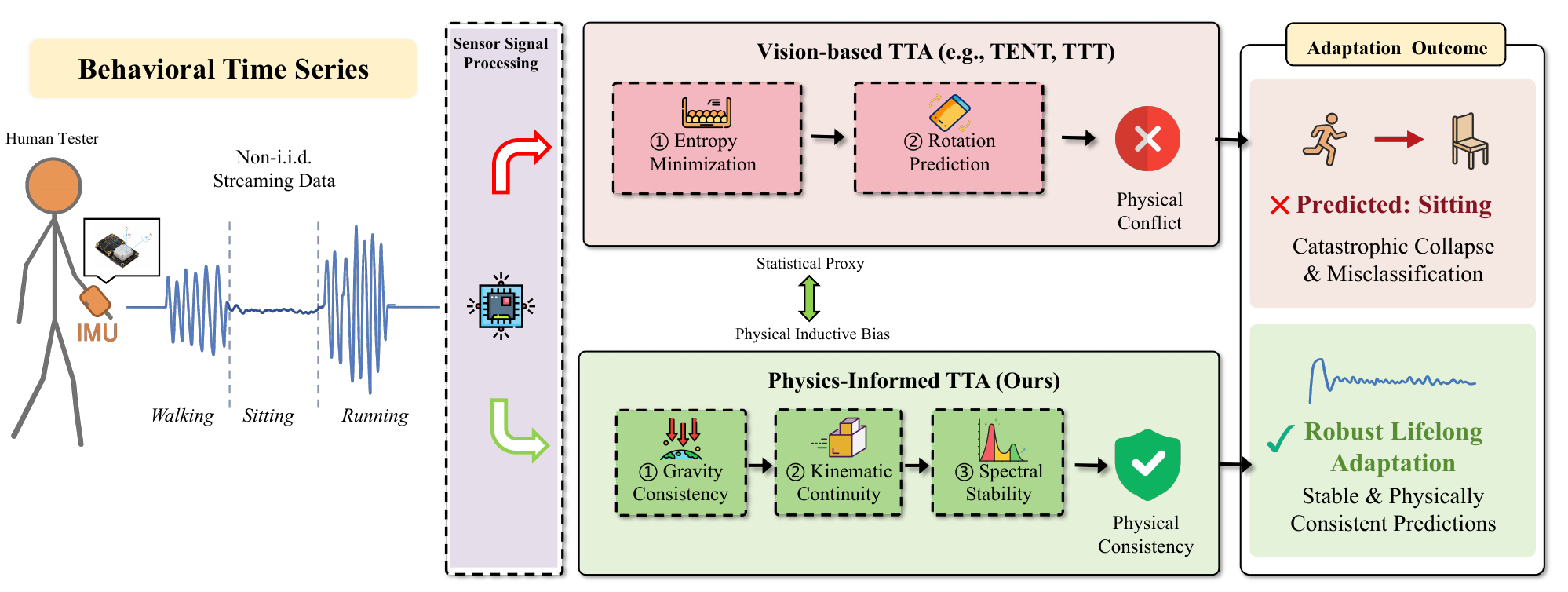}
\caption{Motivation. Under temporally correlated and non-stationary inertial streams, vision-style source-free TTA objectives, including entropy minimization and rotation-based pretext tasks, can become misaligned with the physical structure of the signals, producing overconfident errors, collapse, and forgetting~\cite{wang2021tent,sun2020ttt,gong2022note,zhao2023ttab}. A representative symptom is physical infeasibility. In the gravity-feasibility ribbon, entropy minimization can drive $\|\hat{\mathbf{g}}\|$ far outside the physically plausible band $[0.9\,\mathrm{g},1.1\,\mathrm{g}]$, especially after late-phase transitions, whereas \textsc{PI-TTA} remains stable near $1.0\,\mathrm{g}$ by anchoring online updates with physics-informed stabilization signals.}
\label{fig:motivation}
\end{figure*}

A prominent line of source-free TTA in computer vision updates a small subset of parameters, often normalization affine parameters or running statistics, by minimizing prediction entropy~\cite{wang2021tent,ioffe2015batchnorm,li2016adabn}. Another line performs test-time training with self-supervised auxiliary tasks such as rotation prediction to construct an online learning signal~\cite{sun2020ttt}. While effective in many vision settings, these objectives can become unstable on correlated behavioral streams. Under prolonged single-regime segments, repeated updates may sharpen confidence on the current regime while gradually weakening the representation of others. This can lead to a low-entropy trap in which confidence increases but the adaptation trajectory drifts away from physically plausible structure, causing collapse and forgetting once the stream transitions~\cite{Wang_2022_CVPR_CoTTA,gong2022note,zhang2022memo,niu2022eata,gong2023sotta,zhao2023ttab}. As illustrated in Fig.~\ref{fig:motivation}, confidence-driven adaptation on correlated inertial streams can reduce entropy while drifting away from physically plausible signal structure, eventually producing overconfident errors and post-transition collapse. In mobile deployment, such instability is particularly problematic because adaptation opportunities are limited, recovery is expensive, and sustained execution must remain reliable.

Our key observation is that behavioral inertial streams are not arbitrary unlabeled sequences. They are constrained by physical structure that remains informative across users, activities, and time. Gravity provides a global reference, human motion exhibits short-horizon continuity, and activity dynamics induce characteristic spectral patterns under moderate acquisition variability. These regularities suggest that online adaptation should not be driven by confidence alone. Instead, it should be regularized by lightweight physically grounded signals that remain meaningful under deployment shift. This view is aligned with the broader perspective of physics-informed learning, in which physical structure serves as an inductive bias for robust optimization under non-stationarity~\cite{raissi2019pinn,greydanus2019hnn}.

Motivated by this observation, we propose \textsc{PI-TTA}, a lightweight physics-informed stabilization framework for source-free test-time adaptation on behavioral time series. \textsc{PI-TTA} preserves the standard lightweight adaptation interface used by strong baselines by updating the same small parameter subset, such as normalization affine parameters or running statistics~\cite{wang2021tent,ioffe2015batchnorm,li2016adabn}. On top of this interface, it introduces a physics-informed adaptation objective that combines confidence-driven updates with model-coupled stabilization terms derived from three complementary cues: gravity consistency, short-horizon temporal continuity, and spectral stability under sampling-rate drift. The design aims to stabilize the update trajectory on correlated inertial streams while remaining compatible with practical on-device deployment.

The main contributions of this work are summarized as follows.
\begin{enumerate}
    \item We formulate source-free adaptation for mobile HAR as a deployment-constrained online learning problem on temporally correlated behavioral streams, and analyze why widely used vision-style TTA objectives can become unstable in this setting, leading to overconfident collapse and catastrophic forgetting~\cite{wang2021tent,sun2020ttt,gong2022note,gong2023sotta,zhao2023ttab}.

    \item We introduce \textsc{PI-TTA}, a lightweight physics-informed stabilization framework for source-free test-time adaptation on correlated inertial streams. The method preserves the standard lightweight update scope of normalization-based baselines while incorporating model-coupled adaptation terms with physically grounded stabilization signals.

    \item We establish a comprehensive evaluation suite for source-free adaptation on mobile inertial streams, including long-sequence stress tests, factorized and compound physical shifts, and failure diagnostics. Across USCHAD, PAMAP2, and mHealth, \textsc{PI-TTA} improves long-sequence accuracy by \textbf{9.13\%}, \textbf{4.24\%}, and \textbf{8.82\%}, respectively. It also exposes and mitigates the low-entropy trap: while confidence-driven baselines can produce gravity-proxy magnitudes as extreme as \textbf{13.2\,g}, \textsc{PI-TTA} remains anchored near the \textbf{1.0\,g} feasible band and reduces physical-violation rates by up to \textbf{45.4\%}.

    \item We provide a deployment-oriented characterization of on-device adaptation under practical runtime constraints. Through latency--accuracy Pareto analysis, time-constrained streaming evaluation, and 30-minute sustained execution profiling, we show that \textsc{PI-TTA} improves adaptation stability while remaining compatible with realistic mobile sensing budgets, including operation under a strict 50\,Hz/20\,ms update deadline with an appropriate sparse update schedule.
\end{enumerate}

\section{Related Work}
Our work is most closely related to four directions: sensor-based HAR under deployment shift, source-free test-time adaptation on correlated streams, physics-informed learning for inertial time series, and deployment-constrained on-device adaptation. These directions provide important context, but none directly addresses the setting considered here: \emph{source-free within-session adaptation on strongly correlated mobile HAR streams under strict runtime and memory budgets}. We briefly review each area and clarify the gap addressed by \textsc{PI-TTA}.

\subsection{Sensor-Based HAR Under Deployment Shift}
Deep learning has substantially advanced sensor-based human activity recognition (HAR) and enabled a broad range of mobile and wearable applications~\cite{jiang2015wearablecnn,tang2022tcdattention,haresamudram2020maskedhar}. However, real-world deployment remains challenging because the test distribution often differs from the training condition in ways that are structured, persistent, and behavior-dependent. In wearable inertial sensing, such shifts commonly arise from sensor rotation, body-location changes, subject heterogeneity, acquisition mismatch, and hardware-dependent sampling characteristics~\cite{sztyler2016onbody,chang2020uda_har,qin2019asttl,chen2020metier,bai2020damun,kwon2020imutube}. Among these factors, rotation, placement change, and sampling-rate drift are particularly challenging because they alter temporal dynamics and distort frequency structure even within a single continuous session.

Most prior research on HAR under shift has focused on offline robustness, source-dependent domain adaptation, cross-dataset transfer, or shift-invariant representation learning~\cite{chang2020uda_har,qin2019asttl,bai2020damun,haresamudram2020maskedhar}. These directions improve pre-deployment generalization, but they provide limited guidance for \emph{source-free online adaptation} once a model is already running on-device and must respond to an evolving correlated stream. In this regime, the challenge is not only accuracy under shift, but also whether adaptation remains stable over time, recovers after regime transitions, and fits sustained mobile execution constraints.

\subsection{Source-Free Test-Time Adaptation on Correlated Streams}
Source-free test-time adaptation (TTA) updates a pretrained model at deployment time using only unlabeled target samples, without access to source data. Representative methods adapt a small subset of parameters, often normalization affine parameters or running statistics, by minimizing prediction entropy~\cite{wang2021tent,ioffe2015batchnorm,li2016adabn}, or perform test-time training with auxiliary self-supervised objectives such as rotation prediction~\cite{sun2020ttt}. While these approaches have shown strong results on standard image benchmarks, they can become much less reliable on correlated streams, where successive updates are repeatedly biased by a narrow local regime.

Recent work has made this limitation increasingly explicit. Continual and robust TTA methods have introduced anti-forgetting mechanisms, selective updating, teacher-student smoothing, and entropy-control strategies to mitigate collapse under non-stationary test dynamics~\cite{Wang_2022_CVPR_CoTTA,gong2022note,zhang2022memo,niu2022eata,gong2023sotta,zhao2023ttab,alfarra2024ttatime}. More recent streaming and non-i.i.d.\ variants further emphasize that performance depends on \emph{how} the target stream evolves over time, not only on the marginal shift itself. Representative examples include feature-alignment-based adaptation under dynamic domains~\cite{ye2024datta}, order-aware or temporal-prior strategies that exploit stream ordering~\cite{kim2026oatta}, stateful prototype or classifier-state adaptation~\cite{schirmer2024stad}, diversity-aware buffering to reduce myopic updates~\cite{dobler2024dab}, and adaptive reset mechanisms for recovery from harmful drift~\cite{lim2026asr}.

Nevertheless, the available evidence remains largely image-centric or assumes weaker physical structure than mobile inertial HAR. Moreover, prior stabilization mechanisms are still primarily statistical, including entropy shaping, temporal smoothing, buffering, reset logic, and feature alignment. By contrast, our setting involves inertial streams whose failure modes are closely tied to physical structure, including gravity anchoring, short-horizon motion continuity, and spectral distortion under sampling-rate drift. \textsc{PI-TTA} is therefore complementary to recent streaming TTA methods: instead of relying only on statistical confidence control or stream-state management, it injects lightweight physics-consistent stabilization signals into source-free on-device adaptation for correlated behavioral time series.

\subsection{Physics-Informed Learning for Inertial Signals}
For inertial time series, adaptation instability is amplified by the mismatch between generic statistical objectives and the physical structure of the signals. Unlike visual inputs, inertial measurements are tied to interpretable regularities, including gravity as a global reference, short-horizon motion continuity, and characteristic spectral patterns that remain informative under moderate acquisition variability. These properties suggest that physical consistency can serve as a lightweight and task-relevant anchor for adaptation without labels, source replay, or manually designed target supervision.

More broadly, physics-informed learning has shown that embedding physical structure as an inductive bias can regularize optimization and improve robustness under non-stationarity~\cite{raissi2019pinn,greydanus2019hnn}. However, most prior physics-informed approaches are developed for offline modeling, training-time regularization, or systems with explicit governing equations. They are typically not designed for \emph{source-free online adaptation} on mobile deployment streams, where physical cues must be lightweight, label-free, and compatible with real-time parameter updates. Our use of physics is therefore intentionally pragmatic: rather than solving a full inverse problem, \textsc{PI-TTA} uses simple consistency signals derived from inertial structure to stabilize online adaptation under realistic behavioral shifts.

\subsection{Deployment-Constrained On-Device Adaptation}
Mobile HAR systems operate under tight latency, memory, and energy budgets, making deployability a first-class objective in mobile computing research~\cite{shi2016edge,lim2020fedcomst,li2020flspm,luo2025bideepvit,luo2025activitymamba}. This has motivated efficient backbones, compact temporal models, and deployment-oriented evaluation protocols in which practical utility depends not only on recognition accuracy but also on sustained on-device efficiency. However, efficient inference alone does not guarantee practical adaptation once online updates are introduced. Under correlated non-stationary streams, even lightweight TTA can become unsuitable for deployment if it induces unstable update trajectories, growing state overhead, or adaptation opportunities that exceed the stream-time budget.

Related paradigms, including privacy-preserving personalization and federated optimization, also study adaptation under system constraints~\cite{mcm2017fedavg,kairouz2021flsurvey,bonawitz2017secureagg,li2020fedprox,reddi2021fedopt}. More recent efforts extend efficient adaptation to foundation models through prompt tuning, heterogeneous LoRA allocation, and communication-aware fine-tuning~\cite{che2023fedpeptao,ye2024openfedllm,wang2024flora,bai2024flexlora,cho2024hetlora}. These directions are relevant from a systems perspective, but they mainly address cross-device coordination, communication efficiency, or privacy-preserving personalization. Our setting is more local and more constrained: a single mobile device must adapt to its own evolving unlabeled stream in real time, without source replay and without expensive recovery mechanisms. Accordingly, the central bottleneck here is not communication or distributed optimization, but \emph{stable source-free online adaptation under sustained deployment constraints}.
\begin{figure*}[t]
\centering
\includegraphics[width=0.85\textwidth]{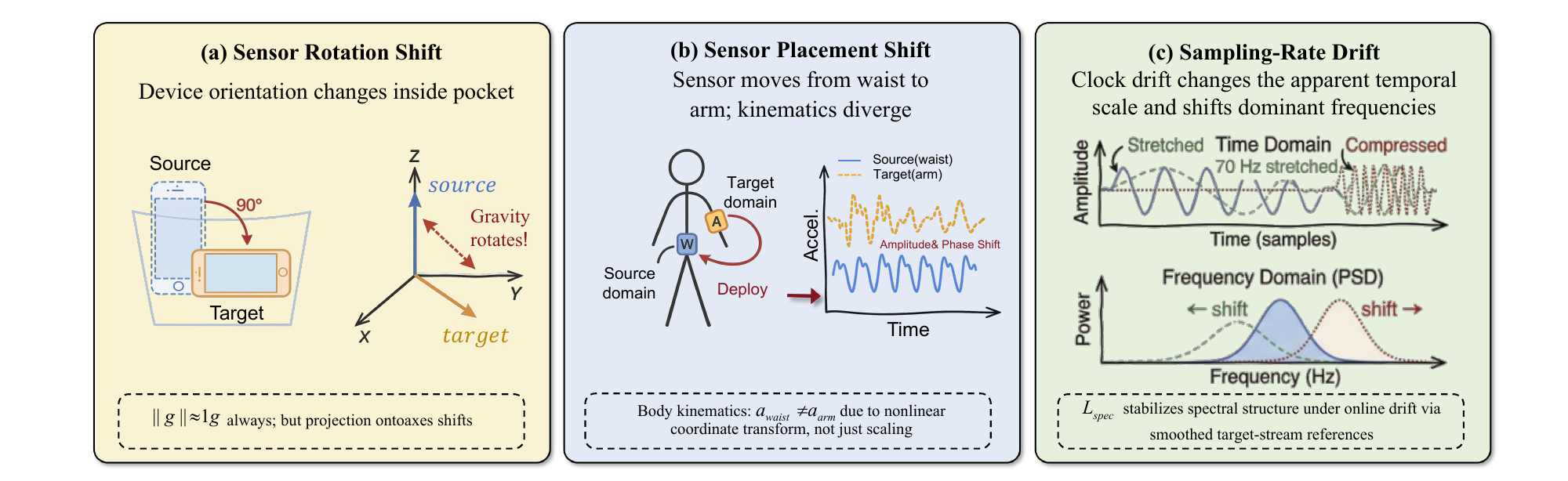}
\caption{Physical shift taxonomy for mobile HAR. We consider three deployment shifts commonly encountered in wearable inertial sensing: sensor rotation ($S_{\text{rot}}$), placement change across body locations ($S_{\text{place}}$), and sampling-rate drift ($S_{\text{drift}}$)~\cite{sztyler2016onbody,chang2020uda_har,qin2019asttl,kwon2020imutube}. These shifts may arise within a single session, alter gravity projection, local motion patterns, and spectral structure, and jointly produce temporally correlated non-i.i.d.\ test streams. This setting motivates stable source-free online adaptation with physics-consistent constraints under mobile resource budgets~\cite{wang2021tent,gong2022note,zhao2023ttab,shi2016edge,alfarra2024ttatime}.}
\label{fig:shift_taxonomy}
\end{figure*}
\section{Why Vision-Style TTA Objectives Become Unstable on Behavioral Streams}
Most vision-style source-free TTA methods were developed and validated under settings with weak temporal correlation. Mobile HAR deployment differs substantially. As summarized in Fig.~\ref{fig:shift_taxonomy}, the test stream is often temporally correlated and non-i.i.d.\ because long single-activity segments are common, while shift factors such as sensor rotation, placement change, and acquisition variability may evolve within a session~\cite{sztyler2016onbody,chang2020uda_har,qin2019asttl,kwon2020imutube}. Under this streaming regime, adaptation objectives based on generic statistical proxies or vision-style geometric pretext tasks can become unstable. We focus on two representative failure mechanisms that motivate the use of physics-consistent adaptation signals.

\subsection{Entropy Minimization: The Low-Entropy Trap and Self-Reinforcing Collapse}
A representative entropy-minimization strategy updates a small subset of parameters, often normalization affine parameters or running statistics, by reducing predictive entropy~\cite{wang2021tent,ioffe2015batchnorm,li2016adabn}:
\begin{equation}
\min_{\theta_{\mathcal{A}}}\ \mathcal{L}_{\text{ent}}
=
\mathbb{E}_{x \sim \mathcal{B}_t}\big[H(\hat{y}(x))\big],
\quad
H(\hat{y})=-\sum_{c}\hat{y}_c\log\hat{y}_c.
\end{equation}
In an i.i.d.\ setting, entropy reduction can encourage confident predictions. On a correlated behavioral stream, however, each mini-batch $\mathcal{B}_t$ may be dominated by a single activity over an extended period. This creates a self-reinforcing failure pattern. First, locally homogeneous batches bias the running normalization statistics toward the current regime. Second, repeated updates sharpen confidence on this local regime while gradually weakening the global class structure.

When the stream transitions to a new activity, the model can fall into a \emph{low-entropy trap}: it produces predictions that are highly confident but incorrect. These overconfident errors then contaminate subsequent updates and further reinforce the drift. Recent continual TTA studies have recognized this instability and proposed statistical anti-forgetting mechanisms~\cite{gong2022note,Wang_2022_CVPR_CoTTA,niu2022eata,gong2023sotta,zhao2023ttab}. In mobile HAR, the problem is even more difficult because limited compute budgets and strict latency constraints make recovery from harmful drift more costly~\cite{alfarra2024ttatime,shi2016edge}.

\subsection{Rotation Pretext Tasks: Mismatch With the Inertial Reference Frame}
An alternative TTA paradigm, test-time training (TTT), obtains an online learning signal by optimizing a self-supervised auxiliary task, most notably rotation prediction~\cite{sun2020ttt}. In computer vision, discrete image rotations such as 0$^\circ$, 90$^\circ$, and 180$^\circ$ can serve as effective synthetic classes because they encourage the network to learn robust spatial representations.

In inertial sensing, however, sensor rotation is not a synthetic augmentation but a continuous physical coordinate transformation. When a mobile device rotates in a pocket, it redistributes the global gravity vector and local body kinematics across the three sensor axes~\cite{sztyler2016onbody,chang2020uda_har}. Treating continuous inertial rotation as a discrete auxiliary task ignores this underlying physical structure. As a result, the network may improve the pretext objective while drifting away from physically plausible states, thereby degrading downstream HAR adaptation. This highlights an important limitation: geometric self-supervision borrowed from vision does not provide a stable universal anchor under realistic inertial shifts. Physical structure should therefore not remain only a diagnostic tool, but should enter the adaptation objective directly as a lightweight stabilizing prior~\cite{raissi2019pinn,greydanus2019hnn}.

\subsection{Design Implications for Stable On-Device Adaptation}
These failure mechanisms lead to three design requirements. First, adaptation needs a stable physical anchor to restore a meaningful reference under orientation shift, rather than relying only on artificial pretext tasks (\emph{gravity consistency}). Second, it must suppress self-reinforcing drift under prolonged single-activity segments (\emph{short-horizon temporal continuity}). Third, it must remain robust to acquisition variability that perturbs temporal and spectral structure (\emph{spectral stability}). Together, these components form a physically grounded, model-coupled stabilization framework for on-device adaptation.
\section{Method: Physics-Informed Test-Time Adaptation (\textsc{PI-TTA})}

\subsection{Overview}
\textsc{PI-TTA} targets source-free adaptation on temporally correlated mobile HAR streams, where the main challenge is not adaptation capacity but update stability. Because the target stream is non-i.i.d.\ and evolves over time, repeated online updates can become dominated by local stream statistics and gradually drive the model toward overconfident but fragile states. Our goal is therefore not to enlarge the adaptation interface, but to stabilize the update trajectory under realistic behavioral shifts.

\begin{figure*}[t]
\centering
\includegraphics[width=0.8\textwidth]{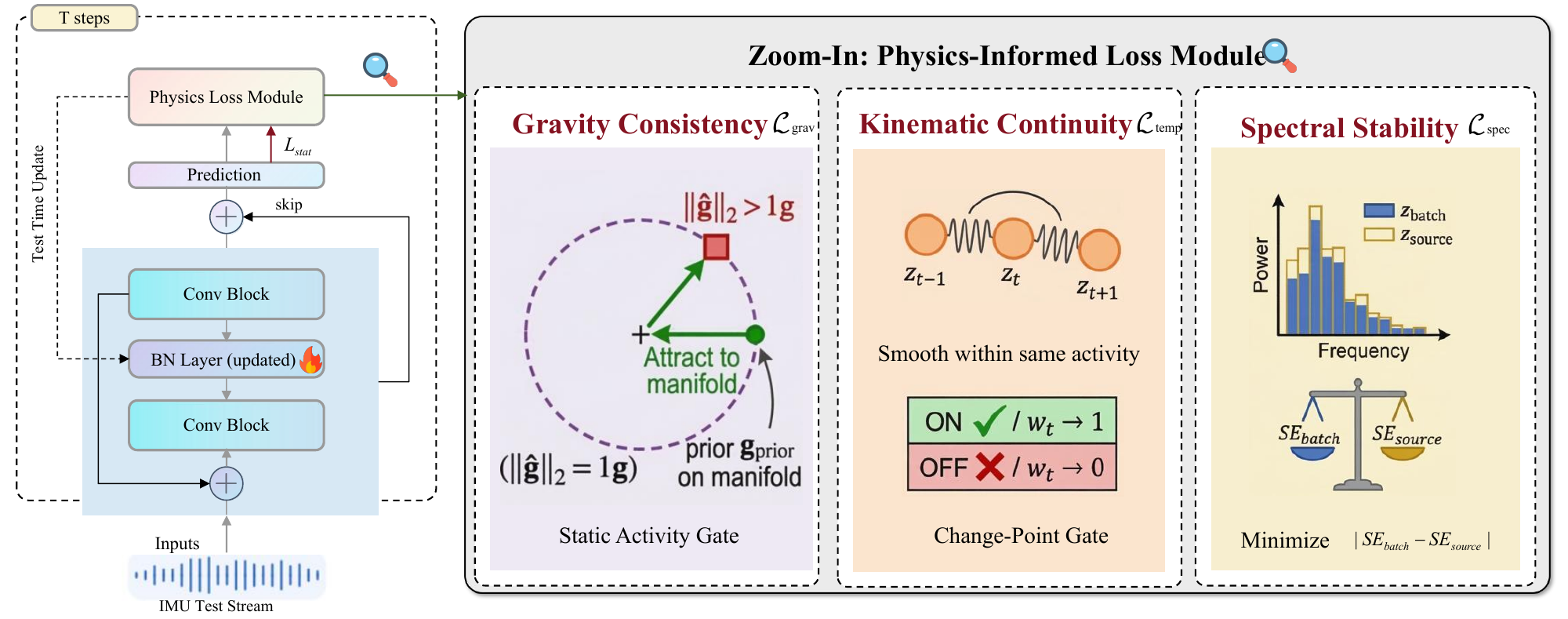}
\caption{\textsc{PI-TTA} overview. Online adaptation updates the same lightweight parameter subset as normalization-based baselines, while gravity- and spectrum-based stabilization terms are computed on model-coupled adapted stream representations. In the zoom-in panel, the three physics-guided components should be labeled \emph{Gravity Consistency}, \emph{Short-Horizon Temporal Continuity}, and \emph{Spectral Stability}.}
\label{fig:method_overview}
\end{figure*}

To preserve fairness and deployability, \textsc{PI-TTA} adopts the same lightweight adaptation interface used by strong source-free baselines, namely a small online parameter subset such as normalization affine parameters and, when enabled, running statistics~\cite{wang2021tent,ioffe2015batchnorm,li2016adabn}. As shown in Fig.~\ref{fig:method_overview}, the method combines a standard statistical adaptation objective with physically grounded stabilization terms and control signals derived from inertial structure.

\subsection{Adaptation Interface and Objective}
Let $f_{\theta}$ denote a pretrained HAR model. We partition its parameters into a frozen subset $\theta_{\mathcal{F}}$ and an adaptable subset $\theta_{\mathcal{A}}$, where $\theta_{\mathcal{A}}$ contains the same lightweight parameters updated by normalization-based baselines. Online adaptation updates only $\theta_{\mathcal{A}}$:
\begin{equation}
\theta = (\theta_{\mathcal{F}}, \theta_{\mathcal{A}}), \qquad
\theta_{\mathcal{A}}^{(t+1)} \leftarrow \theta_{\mathcal{A}}^{(t)} - \eta \nabla_{\theta_{\mathcal{A}}}\mathcal{L}_{\text{PI-TTA}}(\mathcal{B}_t;\theta^{(t)}).
\end{equation}

We define
\begin{equation}
\mathcal{L}_{\text{PI-TTA}}
=
\mathcal{L}_{\text{stat}}
+\lambda_{\text{temp}}\,\mathcal{L}_{\text{temp}}
+\lambda_{\text{grav}}^{(t)}\,\mathcal{L}_{\text{grav}}
+\lambda_{\text{spec}}^{(t)}\,\mathcal{L}_{\text{spec}},
\label{eq:pittta_main}
\end{equation}
where $\mathcal{L}_{\text{stat}}$ is the primary unlabeled adaptation signal, $\mathcal{L}_{\text{temp}}$ is a direct feature-level temporal stabilization term, and $\mathcal{L}_{\text{grav}}$ and $\mathcal{L}_{\text{spec}}$ are gravity- and spectrum-based stabilization terms computed on model-coupled stream representations. Here $\lambda_{\text{temp}}$ is a fixed coefficient controlling the strength of temporal continuity, while $\lambda_{\text{grav}}^{(t)}$ and $\lambda_{\text{spec}}^{(t)}$ are time-varying coefficients that incorporate reliability gating for the gravity and spectral terms, respectively.

In this work, we instantiate $\mathcal{L}_{\text{stat}}$ as entropy minimization, following \textsc{TENT}~\cite{wang2021tent}:
\begin{equation}
\mathcal{L}_{\text{stat}}(\mathcal{B}_t;\theta)
=
\mathbb{E}_{x\sim \mathcal{B}_t}\big[H(\hat{y}(x))\big],
\qquad
H(\hat{y})=-\sum_c \hat{y}_c\log \hat{y}_c.
\end{equation}
Entropy therefore provides the primary update signal, while the remaining terms stabilize the update trajectory.

\subsection{Parameter Coupling and Optimization Path}
A central requirement is that the physical terms influence the adaptable parameters $\theta_{\mathcal{A}}$ through a clear optimization path. To this end, the gravity and spectral terms are not computed from detached raw-input summaries, but from an adapted low-level stream representation
\begin{equation}
\tilde{x}_t=\psi_{\theta_{\mathcal{A}}}(x_t),
\end{equation}
which passes through the same lightweight normalization path governed by $\theta_{\mathcal{A}}$. As a result, $\mathcal{L}_{\text{grav}}$ and $\mathcal{L}_{\text{spec}}$ remain differentiably coupled to $\theta_{\mathcal{A}}$ through $\psi_{\theta_{\mathcal{A}}}(\cdot)$. In practice, reliability-aware modulation is implemented directly through the time-varying coefficients $\lambda_{\text{grav}}^{(t)}$ and $\lambda_{\text{spec}}^{(t)}$ defined in the following subsections; there is no separate third gating mechanism beyond these coefficients and their associated batch-dependent factors.

In our implementation, $\psi_{\theta_{\mathcal{A}}}(\cdot)$ denotes the low-level normalized stream representation immediately after the adaptable normalization path and before deeper feature mixing. This choice preserves local inertial structure and, under the channel-wise normalization instantiation used here, maintains correspondence to the original inertial axes while remaining differentiably coupled to $\theta_{\mathcal{A}}$.

\subsection{Gravity Consistency Signal}
Gravity provides a lightweight physical anchor for inertial adaptation, but batch-level mean acceleration is only a coarse proxy because it can be contaminated by body acceleration. We therefore use a reliability-gated gravity consistency signal rather than assuming exact decomposition.

Let $\tilde{x}_t=\psi_{\theta_{\mathcal{A}}}(x_t)$ denote the adapted stream representation. We compute a low-level adapted inertial summary and maintain a running gravity-proxy estimate:
\begin{equation}
\begin{split}
\bar{\mathbf{a}}_t &= \mathbb{E}_{x\sim \mathcal{B}_t}\big[\tilde{\mathbf{a}}(x)\big],\\
\hat{\mathbf{g}}_t &= \alpha_g \hat{\mathbf{g}}_{t-1} + (1-\alpha_g)\bar{\mathbf{a}}_t.
\end{split}
\label{eq:grav_ema_new}
\end{equation}
When $\psi_{\theta_{\mathcal{A}}}(\cdot)$ is instantiated by channel-wise normalization layers, this adapted representation remains aligned with the original inertial axes up to affine normalization, making it suitable for lightweight gravity-proxy estimation.

We define a feasible magnitude band
\begin{equation}
\mathcal{G}=\{\mathbf{g}\in\mathbb{R}^3: 0.9\,\mathrm{g}\le \|\mathbf{g}\|_2 \le 1.1\,\mathrm{g}\},
\end{equation}
and construct
\begin{equation}
\mathcal{L}_{\text{grav}}
=
\big\|\hat{\mathbf{g}}_t-\Pi_{\mathcal{G}}(\hat{\mathbf{g}}_t)\big\|_2^2
+\lambda_{\text{dir}}\left(1-\left\langle
\frac{\hat{\mathbf{g}}_t}{\|\hat{\mathbf{g}}_t\|_2+\epsilon},
\mathbf{g}_{\text{ref}}
\right\rangle\right),
\label{eq:lgrav_main}
\end{equation}
where $\Pi_{\mathcal{G}}$ denotes Euclidean projection onto the feasible magnitude band. Concretely, for any $\mathbf{u}\in\mathbb{R}^3$,
\begin{equation}
\Pi_{\mathcal{G}}(\mathbf{u})=
\begin{cases}
0.9\,\mathrm{g}\,\dfrac{\mathbf{u}}{\|\mathbf{u}\|_2+\epsilon}, & \|\mathbf{u}\|_2 < 0.9\,\mathrm{g},\\[6pt]
\mathbf{u}, & 0.9\,\mathrm{g}\le \|\mathbf{u}\|_2 \le 1.1\,\mathrm{g},\\[6pt]
1.1\,\mathrm{g}\,\dfrac{\mathbf{u}}{\|\mathbf{u}\|_2+\epsilon}, & \|\mathbf{u}\|_2 > 1.1\,\mathrm{g}.
\end{cases}
\label{eq:proj_g_band}
\end{equation}
The second term in Eq.~\eqref{eq:lgrav_main} discourages slow directional drift relative to a soft EMA reference $\mathbf{g}_{\text{ref}}$, updated by
\begin{equation}
\mathbf{g}_{\text{ref}} \leftarrow \alpha_r \mathbf{g}_{\text{ref}} + (1-\alpha_r)\frac{\hat{\mathbf{g}}_t}{\|\hat{\mathbf{g}}_t\|_2+\epsilon},
\label{eq:gref_update}
\end{equation}
with the update optionally suspended or down-weighted when the batch is judged unreliable. To prevent this term from dominating when the proxy is noisy, we use a variance-based gate
\begin{equation}
\lambda_{\text{grav}}^{(t)}=\lambda_{\text{grav}} \exp(-v_t/\tau_g),
\label{eq:lambda_g}
\end{equation}
where $v_t$ is the within-batch adapted acceleration variance. This time-varying coefficient is the practical reliability-aware gate for the gravity term.

For reporting, $\|\hat{\mathbf g}_t\|$ is mapped back to the input-scale convention used during preprocessing and expressed in gravity-equivalent units for interpretability. This quantity should therefore be interpreted as a proxy scale rather than a calibrated physical accelerometer measurement. Accordingly, the gravity term serves as a practical feasibility and consistency signal rather than an exact physical loss.

\subsection{Short-Horizon Temporal Continuity}
Behavioral dynamics are locally smooth over short horizons, whereas activity boundaries are genuine discontinuities. Let $z_t=\phi_{\theta}(x_t)$ denote the penultimate embedding of window $x_t$. We penalize high-frequency representation variation using
\begin{equation}
\Delta^2 z_t = z_t - 2z_{t-1} + z_{t-2},\qquad
\mathcal{L}_{\text{temp}}=\mathbb{E}_{t}\big[w_t\,\|\Delta^2 z_t\|_2^2\big],
\label{eq:ltemp_new}
\end{equation}
where $w_t\in[0,1]$ is a soft transition-aware gate. Here $c_t=\max_y \hat{y}_t(y)$ denotes prediction confidence, $H_t=H(\hat{y}_t)$ denotes predictive entropy, and $\Delta_y(t)=\mathbb{I}[\arg\max \hat{y}_t \neq \arg\max \hat{y}_{t-1}]$ indicates a label flip. To avoid over-smoothing across genuine activity boundaries, we define
\begin{equation}
w_t = \sigma\!\big(\kappa_c (c_t-\tau_c)\big)\cdot \sigma\!\big(\kappa_h (\tau_h - H_t)\big)\cdot \big(1-\Delta_y(t)\big).
\end{equation}
This is the most direct feature-level stabilization term in \textsc{PI-TTA}, since it acts explicitly on learned representations and therefore has a clear gradient path to $\theta_{\mathcal{A}}$.

\subsection{Spectral Stability Under Sampling-Rate Drift}
Sampling-rate drift distorts temporal frequency structure and can destabilize both statistical adaptation and auxiliary pretext tasks. We therefore regularize spectral statistics computed on the adapted stream representation rather than on frozen raw input.

Let $\tilde{x}_t=\psi_{\theta_{\mathcal{A}}}(x_t)$ and let $\tilde{P}_t(\omega)$ denote its normalized channel-aggregated power spectrum. In practice, $\tilde{P}_t(\omega)$ is obtained by applying a 1-D FFT along the temporal axis of $\psi_{\theta_{\mathcal{A}}}(\mathcal{B}_t)$ channel-wise, followed by channel aggregation and normalization. We maintain an online target-stream reference spectrum
\begin{equation}
P_{\text{ref}} \leftarrow \alpha_s P_{\text{ref}} + (1-\alpha_s)\tilde{P}_t,
\label{eq:pref_update}
\end{equation}
and define
\begin{equation}
M_t=\frac{1}{2}\big(\tilde{P}_t+P_{\text{ref}}\big).
\end{equation}
The Jensen--Shannon component is
\begin{equation}
\mathcal{L}_{\text{JS}}
=
\frac{1}{2}\mathrm{KL}(\tilde{P}_t\|M_t)
+\frac{1}{2}\mathrm{KL}(P_{\text{ref}}\|M_t),
\label{eq:ljs}
\end{equation}
and the spectral-entropy consistency term is
\begin{equation}
\mathcal{L}_{\text{SE}}
=
\big(\mathrm{SE}(\tilde{P}_t)-\mathrm{SE}(P_{\text{ref}})\big)^2,
\end{equation}
where
\begin{equation}
\mathrm{KL}(P\|Q)=\sum_{\omega}P(\omega)\log\frac{P(\omega)+\epsilon}{Q(\omega)+\epsilon},
\end{equation}
and
\begin{equation}
\mathrm{SE}(P)=-\sum_{\omega}P(\omega)\log(P(\omega)+\epsilon).
\end{equation}
The resulting spectral stabilization term is
\begin{equation}
\mathcal{L}_{\text{spec}}
=
\mathcal{L}_{\text{JS}}
+\lambda_{\text{SE}}\mathcal{L}_{\text{SE}},
\label{eq:lspec_new}
\end{equation}
and can optionally be reliability-weighted through
\begin{equation}
\lambda_{\text{spec}}^{(t)}=\lambda_{\text{spec}}\,\gamma_t^{(s)},
\end{equation}
with $\gamma_t^{(s)}\in(0,1]$. This time-varying coefficient is the practical reliability-aware gate for the spectral term. Thus, the spectral term is a model-coupled stabilization term rather than a raw-input drift detector. In implementation, all spectra are computed on discrete frequency bins after channel aggregation and normalization.

\subsection{Online Update Rule, State Maintenance, and Complexity}
At time step $t$, the model receives a batch $\mathcal{B}_t$, computes predictions and embeddings, estimates the model-coupled gravity and spectral quantities, and updates only $\theta_{\mathcal{A}}$ using a small number of SGD steps. All running references are updated online from the target stream, so the procedure remains fully source-free~\cite{alfarra2024ttatime,shi2016edge}.

Algorithm~\ref{alg:pittta} summarizes the online procedure. For window length $T$ and batch size $B$, the dominant additional computation arises from spectral estimation, which costs $O(BT\log T)$ per batch. Exponential moving average reference updates incur only constant-state maintenance, and the extra memory cost is negligible relative to the backbone. \textsc{PI-TTA} requires no replay buffer, no teacher model, and no heavy inner optimization loop, making it compatible with mobile update budgets and sustained on-device execution.

\begin{algorithm}[t]
\caption{Online \textsc{PI-TTA}}
\label{alg:pittta}
\begin{algorithmic}[1]
\STATE \textbf{Input:} pretrained model $f_{\theta_0}$, stream $\{x_t\}$, adaptable subset $\theta_{\mathcal{A}}$, step size $\eta$
\STATE Initialize $\theta \leftarrow \theta_0$; $\hat{\mathbf{g}}_0 \leftarrow \mathbf{0}$; $\mathbf{g}_{\text{ref}}\leftarrow \mathbf{0}$; $P_{\text{ref}} \leftarrow \text{Uniform}$
\FOR{each batch $\mathcal{B}_t$}
  \STATE Forward pass: $\hat{y}_t, z_t \leftarrow f_{\theta}(\mathcal{B}_t)$
  \STATE Compute adapted stream representation $\tilde{x}_t \leftarrow \psi_{\theta_{\mathcal{A}}}(\mathcal{B}_t)$
  \STATE Update $\hat{\mathbf{g}}_t$, $\mathbf{g}_{\text{ref}}$, and $P_{\text{ref}}$ from $\tilde{x}_t$
  \STATE Compute $\mathcal{L}_{\text{PI-TTA}}$ using Eq.~\eqref{eq:pittta_main}
  \STATE Update $\theta_{\mathcal{A}} \leftarrow \theta_{\mathcal{A}} - \eta\nabla_{\theta_{\mathcal{A}}}\mathcal{L}_{\text{PI-TTA}}$
\ENDFOR
\end{algorithmic}
\end{algorithm}
\section{Experimental Setup}

\subsection{Datasets, Preprocessing, and Streaming Construction}
We evaluate on three widely used wearable inertial HAR benchmarks with heterogeneous sensing conditions: USC-HAD (USCHAD)~\cite{zhang2012uschad}, PAMAP2~\cite{reiss2012pamap2}, and mHealth~\cite{banos2014mhealth}. These datasets cover diverse activity sets, sensor placements, and acquisition settings, providing a practical testbed for studying streaming non-stationarity under sensor rotation, placement change, and acquisition drift~\cite{sztyler2016onbody,chang2020uda_har,qin2019asttl,chen2020metier,bai2020damun,kwon2020imutube}. Table~\ref{tab:datasets} summarizes the sensing modality, the primary deployment shifts emphasized for each dataset, and the evaluation protocols in which each dataset is used.

\textbf{Leakage-safe preprocessing.}
Following prior analyses of source-free TTA evaluation pitfalls, we avoid preprocessing that introduces source-domain global statistics into the target stream~\cite{zhao2023ttab}. In particular, we do not apply dataset-level normalization using source means or variances. Instead, all normalization and standardization use only instance-level or current-batch information available at test time, consistent with normalization-based source-free adaptation practice~\cite{ioffe2015batchnorm,li2016adabn,wang2021tent}. Unless otherwise stated, acceleration channels are retained in gravity-equivalent units after sensor-specific calibration so that physically interpretable magnitudes can still be reported. When spectral quantities such as power spectral density (PSD) are required, they are computed online per batch using only target-stream statistics or model-coupled adapted feature streams, depending on the diagnostic or loss definition.

\textbf{Streaming construction.}
All datasets are segmented into fixed-length overlapping windows to emulate realistic on-device streaming inference. Unless otherwise stated, windows are processed strictly in temporal order without shuffling so that temporal correlation and phase transitions are preserved throughout evaluation. This is essential to our problem setting, since the goal is to study source-free adaptation under sustained correlated streams rather than randomized test sampling. Detailed window length, stride, batch size, subject split, and dataset-specific preprocessing choices follow the settings established in prior work~\cite{alfarra2024ttatime}.

\begin{table}[t]
\centering
\caption{Datasets and evaluation usage.}
\label{tab:datasets}
\fontsize{7}{8}\selectfont
\begin{tabular}{lccc}
\toprule
\textbf{Dataset} & \textbf{Modality} & \textbf{Primary shifts} & \textbf{Protocols used} \\
\midrule
USCHAD~\cite{zhang2012uschad} & IMU (acc/gyro) & $S_{\text{rot}}, S_{\text{place}}, S_{\text{drift}}$ & A, B, C \\
PAMAP2~\cite{reiss2012pamap2} & Multi-sensor IMU & $S_{\text{place}}, S_{\text{drift}}$ & A, B, C \\
mHealth~\cite{banos2014mhealth} & IMU + wearable sensors & $S_{\text{place}}, S_{\text{drift}}$ & A, B, C \\
\bottomrule
\end{tabular}
\end{table}

\subsection{Baselines, Update Scope, and Fairness Protocol}
We focus on \emph{source-free} online adaptation methods that update only a lightweight parameter subset at test time, which is a deployment-relevant setting for mobile sensing~\cite{shi2016edge,lim2020fedcomst}. Our main baselines are \textsc{Source Only} (no adaptation), \textsc{TENT} (entropy minimization with normalization-layer updates)~\cite{wang2021tent}, and \textsc{TTT} (rotation-based self-supervision)~\cite{sun2020ttt}. To strengthen the comparison under temporally correlated streams, we additionally include representative continual and robust TTA methods, including \textsc{CoTTA}~\cite{Wang_2022_CVPR_CoTTA}, \textsc{NOTE}~\cite{gong2022note}, \textsc{EATA}~\cite{niu2022eata}, \textsc{SoTTA}~\cite{gong2023sotta}, \textsc{MEMO}~\cite{zhang2022memo}, and \textsc{SAR}~\cite{niu2023sar}. For each method, we follow the original update rule as closely as possible under a unified source-free streaming protocol.

These baselines span two broad categories. The first includes confidence-driven or self-supervision-driven methods, such as \textsc{TENT} and \textsc{TTT}. The second includes continual or anti-forgetting methods, such as \textsc{CoTTA}, \textsc{NOTE}, \textsc{EATA}, and \textsc{SAR}. This grouping clarifies the role of \textsc{PI-TTA}: the goal is not merely to add another stabilization heuristic, but to test whether physics-informed adaptation remains stable even when stronger generic retention mechanisms are already present.

\textbf{Update scope and fairness.}
Unless a baseline intrinsically requires additional modules, such as the rotation head in \textsc{TTT}, we restrict online updates to normalization affine parameters and, where applicable, running statistics. This choice follows standard source-free TTA practice~\cite{ioffe2015batchnorm,li2016adabn,wang2021tent} and ensures that performance differences are not driven by substantially different adaptation capacity. Our method uses the same lightweight update scope as normalization-based baselines and differs primarily in the stabilization objective. All adaptive baselines are evaluated under the same streaming order and the same batch schedule. Unless a protocol explicitly studies sparse updating, methods are updated at every batch; when an update interval $K>1$ is introduced, the same interval is applied across compared methods within that protocol. Methods such as OATTA, CoTTA, NOTE, and SAR are run under their closest reproducible settings while preserving their original auxiliary states or memory mechanisms when these are intrinsic to the method. When architectural differences prevent exactly identical parameterization, we use comparable update scope whenever applicable and keep the stream schedule, runtime setting, and reporting protocol fixed.

\textbf{Representative deployment cost.}
Table~\ref{tab:baselines_cost} summarizes the update scope, the main stabilization signal, and representative per-step deployment overhead under a shared runtime setting. The table is intended to position the compared methods along an adaptation-capacity versus deployability spectrum, rather than to enumerate all compared baselines or replace the more detailed profiling reported later.

\begin{table*}[t]
\centering
\caption{Representative baselines and update characteristics.}
\label{tab:baselines_cost}
\fontsize{8}{9}\selectfont
\setlength{\tabcolsep}{5pt}
\begin{tabular}{lccccc}
\toprule
\textbf{Method} & \textbf{Update scope} & \textbf{Main stabilization signal} & \textbf{Latency (ms)} & \textbf{Peak memory (MB)} & \textbf{Energy (mJ)} \\
\midrule
Source Only & Inference only & None & 15.2 & 18.5 & 8.5 \\
TENT~\cite{wang2021tent} & BN affine / stats & Entropy minimization & 38.5 & \textbf{26.0} & \textbf{24.0} \\
TTT~\cite{sun2020ttt} & Encoder + rotation head & Geometric pretext & 82.4 & 58.2 & 68.2 \\
EATA~\cite{niu2022eata} & Lightweight subset & Anti-forgetting filter & \textbf{32.4} & 28.5 & 32.4 \\
CoTTA~\cite{Wang_2022_CVPR_CoTTA} & Lightweight subset & Teacher consistency & 65.5 & 75.4 & 68.2 \\
NOTE~\cite{gong2022note} & Lightweight subset & Buffering / debiasing & 38.5 & 45.5 & 38.5 \\
\textsc{PI-TTA} (ours) & BN affine / stats & Physics-consistent regularization & 45.1 & 30.4 & 28.2 \\
\bottomrule
\end{tabular}
\end{table*}

\subsection{Diagnostic Definitions and Measurement Methodology}
Because several analyses go beyond scalar accuracy, we explicitly define the diagnostics and deployment measurements used throughout the paper.

\textbf{Gravity-related diagnostics.}
The quantity used in the gravity-feasibility ribbon and violation analysis is a \emph{model-coupled gravity proxy} rather than a raw batch mean of acceleration. Concretely, it is computed along the same lightweight adapted normalization path used during online updating and is reported in gravity-equivalent units for interpretability. Accordingly, large deviations should be interpreted as evidence of physically implausible adapted signal states, not as direct sensor-certified measurements of rigid-body gravity. The corresponding violation diagnostic checks whether the proxy magnitude falls outside the feasible band $[0.9\,\mathrm{g},1.1\,\mathrm{g}]$.

\textbf{Spectral diagnostics.}
Spectral stability is analyzed using batch-level PSD statistics and spectral entropy. For diagnostics, PSD is computed on the same adapted stream representation used in the spectral term so that the reported quantity remains coupled to the online adaptation path rather than a fixed raw-input summary. This avoids conflating adaptation effects with purely exogenous input statistics.

\textbf{Deployment measurements.}
Latency is measured end-to-end per batch, including forward inference, loss computation, backward update, and parameter write-back when adaptation is enabled. Peak memory is recorded under the same shared runtime setting used in deployment profiling. Energy and temperature are measured on the same device under the same thread, batch-size, and stream-schedule configuration for all methods. Reported values are averaged over repeated runs after warm-up. In our implementation, energy is recorded with the device-side power profiling toolchain, temperature is read from the system thermal interface, warm-up is performed before measurement, and each reported value is averaged over repeated runs with background tasks minimized as much as possible. These measurements are also used in the sustained execution and time-constrained evaluations reported later. Detailed toolchain information, sampling interval, and device configuration are provided in the implementation notes and deployment subsection.

\subsection{Evaluation Protocols}
We design the evaluation suite to probe three properties that are central to source-free adaptation on mobile inertial streams: stability under long-range temporal correlation, robustness under physically meaningful deployment shifts, and interpretability of adaptation failures beyond scalar accuracy~\cite{zhao2023ttab}. Accordingly, the protocols are designed to answer three questions: whether a method remains stable, under which shift regimes it remains effective, and why it fails when it does not. As indicated in Table~\ref{tab:datasets}, all three datasets are used across Protocols A--C so that long-sequence stability, factorized and compound shifts, and failure diagnostics can be examined within a consistent evaluation framework.

\subsubsection{Long-Sequence Stability Under Temporal Correlation}
To stress-test catastrophic forgetting under extreme temporal correlation, we replace the conventional shuffled test stream with a class-sorted stream composed of long contiguous phases, each dominated by a single activity mode. Unless otherwise stated, each phase contains 1,000 streaming steps, producing a strongly non-i.i.d.\ test process that amplifies self-reinforcing adaptation errors.

We report two complementary metrics. \emph{Online Accuracy} is measured on each incoming batch to reflect immediate adaptation behavior under local stream statistics. \emph{Held-out Accuracy} is evaluated every 100 steps on a fixed all-class hold-out set that is never used for adaptation, in order to quantify catastrophic forgetting and loss of global discriminative structure. For datasets such as mHealth, some later phases can be comparatively simpler or lower-diversity local regimes. Accordingly, phase-wise online accuracy in a late segment should not be interpreted in isolation as evidence of stable global adaptation. Unless otherwise specified, all quantitative results are averaged over $n=5$ independent runs and reported as mean $\pm$ standard deviation.

\subsubsection{Shift Breakdown and Compound Physical Shifts}
To identify which adaptation mechanisms remain effective under different deployment perturbations, we evaluate physically grounded shift factors at two levels. We first consider \emph{factorized shifts}, where each perturbation is applied in isolation. We then consider \emph{compound shifts}, where multiple perturbations are introduced sequentially within the same stream to approximate realistic deployment conditions.

For the factorized setting, we evaluate three shift factors: \emph{sensor rotation} $S_{\text{rot}}$, implemented by realistic device-frame rotations of 30$^\circ$, 60$^\circ$, and 90$^\circ$; \emph{placement change} $S_{\text{place}}$, where the sensor location changes across body positions such as waist $\rightarrow$ arm or waist $\rightarrow$ chest; and \emph{sampling-rate drift} $S_{\text{drift}}$, implemented via time-axis warping to emulate acquisition mismatch such as 100\,Hz $\rightarrow$ 70/80/120/130\,Hz.

This factorized design aligns each shift type with a corresponding physical prior in \textsc{PI-TTA}: gravity consistency is most relevant to rotation, short-horizon temporal continuity to correlated transition dynamics, and spectral stability to sampling-rate mismatch. For compound streams, multiple perturbations are accumulated sequentially within a single evaluation run to test whether adaptation remains stable as mismatch factors compound over time.

\subsubsection{Failure Visualization and Diagnostics}
Scalar accuracy alone is insufficient to explain why online adaptation succeeds or fails on behavioral streams. We therefore introduce diagnostics that connect statistical confidence to physical plausibility and representation geometry. Representative diagnostic results are presented in the corresponding result figures, where they are used to visualize the low-entropy trap, gravity infeasibility, and progressive feature collapse under unstable adaptation.

Specifically, we use three complementary diagnostics. First, the \emph{gravity-feasibility ribbon} plots the gravity-proxy magnitude over time together with a feasible band $[0.9\,\mathrm{g},1.1\,\mathrm{g}]$. Second, \emph{t-SNE evolution} visualizes feature embeddings at multiple time checkpoints, while the \emph{silhouette score} quantifies class separability and detects progressive feature collapse. Third, \emph{entropy--physics trade-off plots} jointly track predictive entropy and a physical-violation score, exposing cases in which entropy decreases even as physical inconsistency increases. This mismatch, referred to as the \emph{low-entropy trap}, indicates that the model is becoming more confident for the wrong reasons~\cite{gong2022note,gong2023sotta,zhao2023ttab}.

\subsection{Reporting Conventions}
To ensure comparability across methods and protocols, we adopt unified conventions for optimization, fairness control, deployment profiling, and reproducibility.

\textbf{Optimization.}
We use low-learning-rate SGD for test-time updates to reduce overshoot risk under non-stationary streams~\cite{wang2021tent,zhao2023ttab}. All methods operate fully online without access to source data. To assess robustness under practical tuning uncertainty, we additionally evaluate a shared learning-rate grid $\{10^{-4},10^{-3},10^{-2}\}$ for all adaptive baselines under the same long-sequence streaming protocol.

\textbf{Fairness and reporting.}
All adaptive baselines are evaluated under the same streaming order, the same adaptation trigger frequency, and matched reporting conditions for memory and energy overhead. When architectural differences prevent exactly identical parameterization, we use comparable update scope whenever applicable and control reporting through matched memory, energy, and stream-schedule conditions. We also apply a shared update-interval sweep under the same streaming protocol to reduce the possibility that gains arise from selectively tuned adaptation frequency rather than from the adaptation objective itself.

\textbf{Time-constrained evaluation.}
For streaming scenarios with strict real-time budgets, we additionally use a time-constrained scoring rule. Let $T_{\text{adapt}}$ denote the measured adaptation latency per update and let $T_{\text{budget}}$ denote the inter-arrival time budget implied by the sensor rate. If $T_{\text{adapt}} \leq T_{\text{budget}}$, the update is applied before the next window and is marked \emph{safe}. If $T_{\text{budget}} < T_{\text{adapt}} \leq 2T_{\text{budget}}$, the update is \emph{delayed by one cycle}. If $T_{\text{adapt}} > 2T_{\text{budget}}$, one or more adaptation opportunities are missed and the update is marked \emph{dropped}. Effective accuracy is measured under the resulting delayed-update schedule rather than under an idealized synchronous assumption.

\textbf{Deployment profiling.}
For deployment relevance, we report per-step runtime, memory, and energy overhead, together with a 30-minute sustained execution test on Snapdragon~8~Gen~2 under a CPU single-thread setting. In addition to raw runtime, we also report update-interval studies to expose the effective operating region under constrained adaptation opportunities. These measurements complement the accuracy results by quantifying sustained deployability under edge-runtime constraints.

\textbf{Reproducibility.}
We keep the streaming order intact without shuffling, report both online and held-out metrics, and specify all protocol settings explicitly. Unless otherwise noted, each experiment is repeated over $n=5$ independent runs and reported as mean $\pm$ standard deviation. Detailed implementation choices, dataset-specific preprocessing, hardware profiling settings, and measurement toolchains are reported in the corresponding implementation and deployment sections.
\section{Results}
\subsection{Long-Sequence Stability Under Temporal Correlation}

\begin{figure*}[t]
\centering
\includegraphics[width=0.9\textwidth]{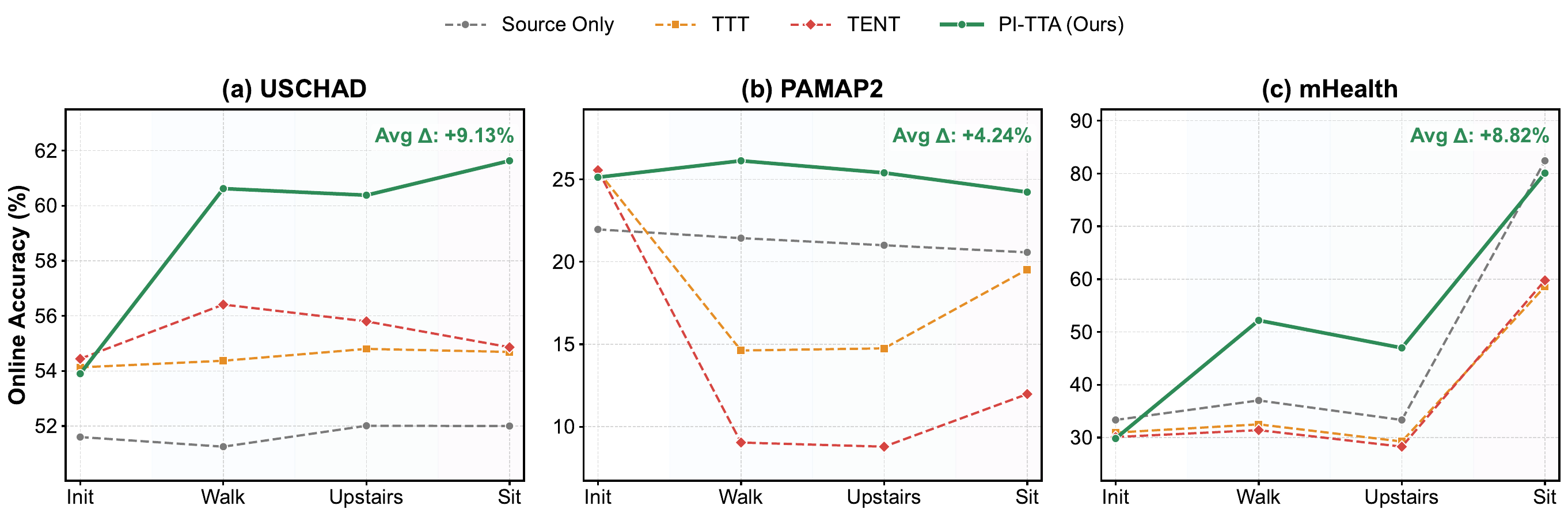}
\caption{Long-sequence class-sorted streaming evaluation under Protocol~A.}
\label{fig:torture}
\end{figure*}

We begin with the primary setting of this work, namely long correlated streams with abrupt regime transitions. Following Protocol~A, we construct a class-sorted stream in which each activity phase contains 1,000 consecutive online steps before switching to the next class. This protocol stresses update stability under prolonged temporal correlation rather than shuffled evaluation. All adaptive methods are evaluated under the same streaming protocol, matched adaptation frequency, comparable update scope whenever applicable, and matched reporting conditions for memory and energy overhead. As illustrated in Fig.~\ref{fig:torture}, this setting creates extended single-regime segments followed by abrupt transitions, making confidence-driven updates particularly vulnerable to self-reinforcing drift.

Table~\ref{tab:torture_main} reports the primary comparison on USCHAD. Each phase contains 1,000 consecutive online steps from a single activity class, and all results are reported as mean $\pm$ standard deviation over $n=5$ runs. Lower values in peak gravity-proxy magnitude, peak memory, and energy indicate better physical plausibility and lower deployment cost. Source Only is included as the non-adaptive deployment reference, and OATTA is included as a representative order-aware streaming baseline.

Four patterns are clear. First, entropy-driven adaptation is brittle under long correlated streams. TENT improves sharply in Phase~1 but collapses after the phase transition, dropping to 18.5\% in Phase~2 and 12.4\% in Phase~3. Second, stronger continual and streaming-aware baselines improve retention, but at noticeably higher system cost. CoTTA, NOTE, SAR, and the order-aware OATTA baseline are all more stable than TENT after regime transitions, yet they still incur substantially larger memory or energy overhead than inference-only execution. Third, \textsc{PI-TTA} remains stable across all three phases, achieving the best Phase~2 and Phase~3 accuracies while keeping the peak gravity-proxy magnitude close to the feasible regime with only modest overhead. Finally, the gains of \textsc{PI-TTA} are not explained simply by adding another generic robust TTA heuristic; they persist even against modern continual, selective-update, and order-aware streaming baselines.

\begin{table*}[t]
\centering
\caption{Primary long-sequence comparison on USCHAD under Protocol~A.}
\label{tab:torture_main}
\fontsize{8}{9}\selectfont
\setlength{\tabcolsep}{5pt}
\begin{tabular}{lcccccc}
\toprule
Method & Phase 1 (Walk) & Phase 2 (Upstairs) & Phase 3 (Sit) & Peak gravity-proxy magnitude $\|\hat{\mathbf g}_t\|$ & Peak Mem (MB) & Energy (mJ/batch) \\
\midrule
Source Only & 51.6$\pm$0.8 & 52.0$\pm$1.0 & 51.5$\pm$0.9 & \textbf{1.01\,g} & \textbf{18.5} & \textbf{8.5} \\
TENT~\cite{wang2021tent} & \textbf{61.2$\pm$1.1} & 18.5$\pm$2.3 & 12.4$\pm$1.5 & 13.20\,g & 26.0 & 24.0 \\
CoTTA~\cite{Wang_2022_CVPR_CoTTA} & 58.5$\pm$0.5 & 48.6$\pm$1.2 & 45.3$\pm$1.5 & 3.20\,g & 75.4 & 68.2 \\
NOTE~\cite{gong2022note} & 60.1$\pm$0.8 & 54.2$\pm$1.4 & 52.8$\pm$1.1 & 2.50\,g & 42.1 & 38.5 \\
SAR~\cite{niu2023sar} & 60.5$\pm$0.9 & 49.8$\pm$1.5 & 46.2$\pm$1.8 & 3.85\,g & 48.5 & 55.4 \\
OATTA~\cite{kim2026oatta} & 60.8$\pm$0.6 & 53.4$\pm$1.1 & 48.5$\pm$1.4 & 2.65\,g & 62.4 & 45.2 \\
\midrule
\textsc{PI-TTA} (Ours) & 60.6$\pm$0.7 & \textbf{60.3$\pm$0.9} & \textbf{61.6$\pm$0.7} & 1.03\,g & 30.4 & 28.2 \\
\bottomrule
\end{tabular}
\end{table*}

Phase-wise online accuracy measures local adaptation quality under the current stream regime, but it does not by itself distinguish stable adaptation from forgetting. A method can perform well on the current phase while gradually degrading the global class structure. We therefore also track held-out accuracy on a fixed all-class set that is never used for adaptation. This isolates retention of the global decision boundary from local in-phase fitting.

As shown in Table~\ref{tab:heldout_forgetting}, TENT progressively destroys global retention, with held-out accuracy dropping from 58.5\% at initialization to 8.2\% by the end of the sequence. CoTTA slows this degradation, but still drifts downward. OATTA provides stronger retention than CoTTA, which is consistent with its explicit use of stream ordering, yet it still loses global structure over the full 3,000-step sequence. By contrast, \textsc{PI-TTA} preserves the held-out decision boundary throughout the full stream and even yields a slight gain over initialization, despite using no replay buffer and no source data.

\begin{table}[t]
\centering
\caption{Held-out accuracy evolution under the USCHAD long-sequence protocol.}
\label{tab:heldout_forgetting}
\fontsize{7}{8}\selectfont
\setlength{\tabcolsep}{4pt}
\resizebox{\linewidth}{!}{
\begin{tabular}{lccccc}
\toprule
\textbf{Phase End} & \textbf{Source Only} & \textbf{\textsc{TENT}} & \textbf{\textsc{CoTTA}} & \textbf{\textsc{OATTA}} & \textbf{\textsc{PI-TTA}} \\
\midrule
$T_0$ (Initial) & 58.5 & 58.5 & 58.5 & 58.5 & 58.5 \\
$T_1$ (End of Walk, Step 1000) & 58.5 & 32.1 & 56.2 & 58.1 & \textbf{59.4} \\
$T_2$ (End of Upstairs, Step 2000) & 58.5 & 15.4 & 52.1 & 54.6 & \textbf{58.8} \\
$T_3$ (End of Sit, Step 3000) & 58.5 & 8.2 & 48.5 & 50.2 & \textbf{59.1} \\
\bottomrule
\end{tabular}}
\end{table}

Held-out samples are never used for online adaptation, so Table~\ref{tab:heldout_forgetting} measures retention of the global decision boundary rather than local in-phase fit.

To examine whether this trend generalizes beyond a single benchmark, Table~\ref{tab:torture_summary} summarizes long-sequence stability on three datasets. The reported values are phase-wise online accuracy under the current local stream regime, and ``Avg.\ $\Delta$'' denotes the average change relative to the initial checkpoint. Because phase-wise online accuracy can be influenced by local class composition, especially in later phases, we interpret it together with held-out retention rather than in isolation. In particular, for mHealth the final phase is comparatively simple, so Phase~3 online accuracy should be interpreted together with held-out retention rather than as a standalone indicator of global stability.

Across USCHAD, PAMAP2, and mHealth, \textsc{PI-TTA} is the only method that consistently preserves positive long-stream gains relative to the initial checkpoint. It improves by \textbf{+9.13}\% on USCHAD, \textbf{+4.24}\% on PAMAP2, and \textbf{+8.82}\% on mHealth, corresponding to an average gain of \textbf{+7.4}\%.

\begin{table}[t]
\centering
\caption{Long-sequence stability summary on three datasets.}
\label{tab:torture_summary}
\fontsize{7}{8}\selectfont
\begin{tabular}{lcccc|c}
\toprule
\multirow{2}{*}{Method} & Initial & Phase 1 & Phase 2 & Phase 3 & Avg.\ $\Delta$ \\
&  & (Walk) & (Upstairs) & (Sit) &  \\
\midrule
\multicolumn{6}{c}{\textbf{USCHAD}~\cite{zhang2012uschad}}\\
Source Only & 51.60 & 51.25 & 52.01 & 52.00 & -- \\
TTT~\cite{sun2020ttt} & 54.13 & 54.37 & 54.80 & 54.69 & +2.87 \\
TENT~\cite{wang2021tent} & 54.44 & 56.41 & 55.80 & 54.86 & +3.94 \\
\textsc{PI-TTA} & 53.90 & \textbf{60.62} & \textbf{60.38} & \textbf{61.63} & \textbf{+9.13} \\
\midrule
\multicolumn{6}{c}{\textbf{PAMAP2}~\cite{reiss2012pamap2}}\\
Source Only & 21.96 & 21.43 & 21.00 & 20.57 & -- \\
TTT~\cite{sun2020ttt} & 25.50 & 14.62 & 14.75 & 19.53 & -4.70 \\
TENT~\cite{wang2021tent} & 25.55 & 9.04 & 8.79 & 11.98 & -11.06 \\
\textsc{PI-TTA} & 25.12 & \textbf{26.12} & \textbf{25.39} & \textbf{24.22} & \textbf{+4.24} \\
\midrule
\multicolumn{6}{c}{\textbf{mHealth}~\cite{banos2014mhealth}}\\
Source Only & 33.33 & 37.03 & 33.33 & \textbf{82.42} & -- \\
TTT~\cite{sun2020ttt} & 30.94 & 32.50 & 29.25 & 58.59 & -10.81 \\
TENT~\cite{wang2021tent} & 30.10 & 31.41 & 28.27 & 59.77 & -11.11 \\
\textsc{PI-TTA} & 29.82 & \textbf{52.19} & \textbf{46.98} & 80.08 & \textbf{+8.82} \\
\bottomrule
\end{tabular}
\end{table}

Overall, \textsc{PI-TTA} delivers the strongest long-sequence stability by improving both local online adaptation and held-out retention under prolonged temporal correlation.

\subsection{Shift Breakdown and Compound Shifts: Factorized Mechanisms and Practical Regime Boundaries}

\begin{figure*}[t]
\centering
\includegraphics[width=0.85\textwidth]{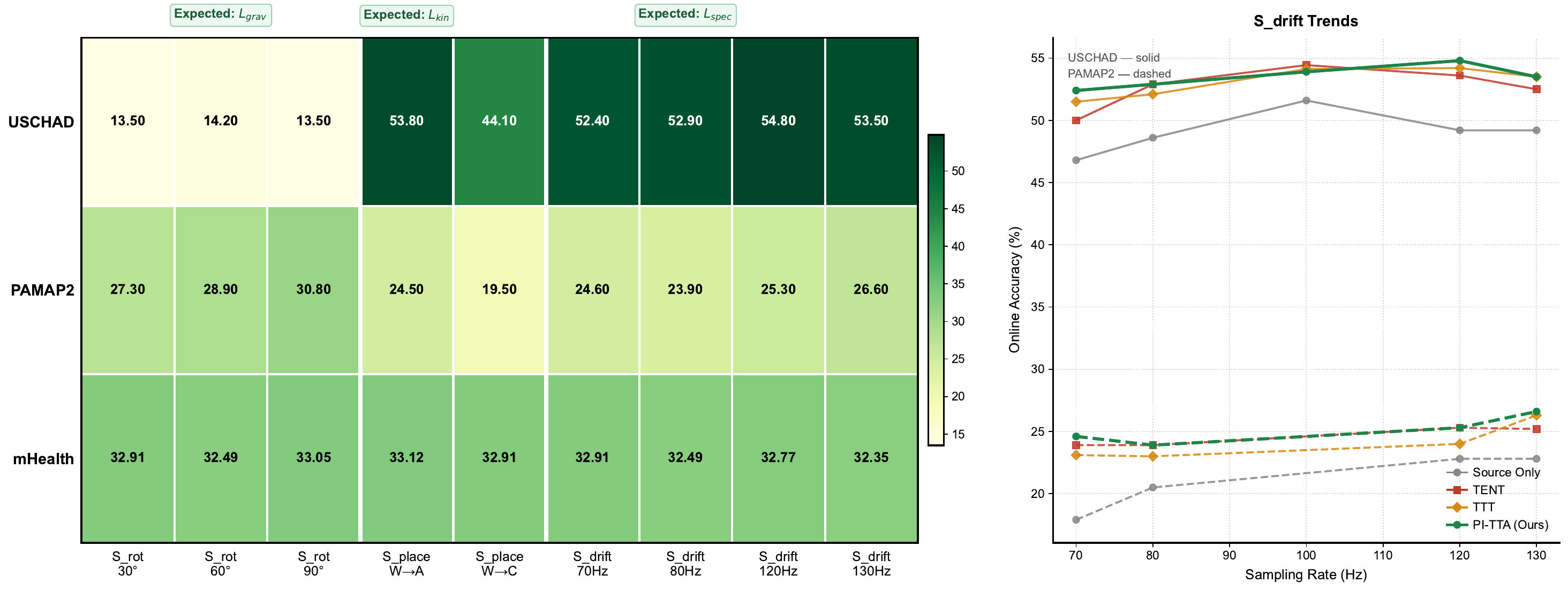}
\caption{Shift breakdown across rotation, placement change, and sampling-rate drift.}
\label{fig:shift_breakdown}
\end{figure*}

We next evaluate robustness under deployment shifts from two complementary perspectives. We first consider \emph{factorized physical shifts}, where rotation, placement change, and sampling-rate drift are examined in isolation. We then consider \emph{compound shift streams}, where multiple perturbations accumulate within the same stream to better approximate realistic mobile deployment conditions~\cite{sztyler2016onbody,chang2020uda_har,kwon2020imutube,zhao2023ttab}. Fig.~\ref{fig:shift_breakdown} reports the factorized comparison.

Fig.~\ref{fig:shift_breakdown} shows that the advantage of \textsc{PI-TTA} is structured rather than uniform. It is strongest when the perturbation directly disrupts physical consistency in orientation, motion continuity, or temporal frequency structure.

\textbf{Rotation ($S_{\text{rot}}$).}
Across 30$^\circ$, 60$^\circ$, and 90$^\circ$ rotations, \textsc{PI-TTA} is best or near-best in most settings, supporting the role of gravity consistency in restoring a valid inertial reference.

\textbf{Placement change ($S_{\text{place}}$).}
\textsc{PI-TTA} performs best in most waist-to-arm and waist-to-chest transfers. Placement change alters not only orientation, but also motion amplitude, inter-axis coupling, and the effective biomechanical trajectory observed by the sensor. This makes transfer difficult for objectives that rely only on generic confidence or simple geometric pretext tasks.

\textbf{Sampling-rate drift ($S_{\text{drift}}$).}
Under mild and moderate drift, including 70--80\,Hz and 120\,Hz, \textsc{PI-TTA} achieves the strongest accuracy in most settings. Under more extreme drift, \textsc{TTT} can approach \textsc{PI-TTA} in some cases, suggesting that geometric self-supervision and spectral stabilization may provide partially complementary signals near severe temporal distortion~\cite{zhao2023ttab}.

Real deployment rarely presents one perturbation at a time. We therefore evaluate a compound shift stream that progressively accumulates multiple mismatches within the same run. Specifically, the stream follows the sequence $T_0 \rightarrow T_1 \rightarrow T_2 \rightarrow T_3$, where $T_0$ is the unshifted baseline condition, $T_1$ introduces a 60$^\circ$ device rotation, $T_2$ adds a placement transfer from waist to arm, and $T_3$ further introduces sampling-rate drift. Table~\ref{tab:compound_shift} summarizes this setting, and the reported spectral entropy serves as a diagnostic of end-stage spectral stability.

\begin{table}[t]
\centering
\caption{Compound-shift robustness under progressively accumulated perturbations.}
\label{tab:compound_shift}
\fontsize{7}{8}\selectfont
\setlength{\tabcolsep}{4pt}
\begin{tabular}{lccccc}
\toprule
Method & $T_0$ Acc & $T_1$ Acc & $T_2$ Acc & $T_3$ Acc & $T_3$ Spectral Entropy \\
\midrule
Source Only & 85.2 & 42.1 & 28.5 & 15.2 & 0.95 \\
TENT~\cite{wang2021tent} & \textbf{86.5} & 45.3 & 18.2 & 10.4 & 0.21 \\
NOTE~\cite{gong2022note} & 86.1 & 68.4 & 52.1 & 38.6 & 1.85 \\
\textsc{PI-TTA} & 86.3 & \textbf{81.5} & \textbf{74.2} & \textbf{71.8} & \textbf{0.98} \\
\bottomrule
\end{tabular}
\end{table}

Table~\ref{tab:compound_shift} shows that the compound setting is substantially more demanding than any factorized shift in isolation. TENT deteriorates sharply once the stream moves beyond pure rotation and becomes particularly unstable after placement transfer is introduced. NOTE remains more robust than TENT in the intermediate stages, but its performance still drops substantially at $T_3$, where sampling-rate drift is added. By contrast, \textsc{PI-TTA} maintains strong accuracy across the entire progression and preserves near-stable spectral behavior at $T_3$. This indicates that the three physical priors act complementarily when perturbations accumulate.

Overall, these results show that \textsc{PI-TTA} provides structured robustness under mixed physical shifts rather than only higher average accuracy.

\subsection{Ablation and Mechanism Validation}

\begin{figure}[t]
\centering
\subfloat[Ablation: per-constraint contribution]{%
  \includegraphics[width=0.82\linewidth]{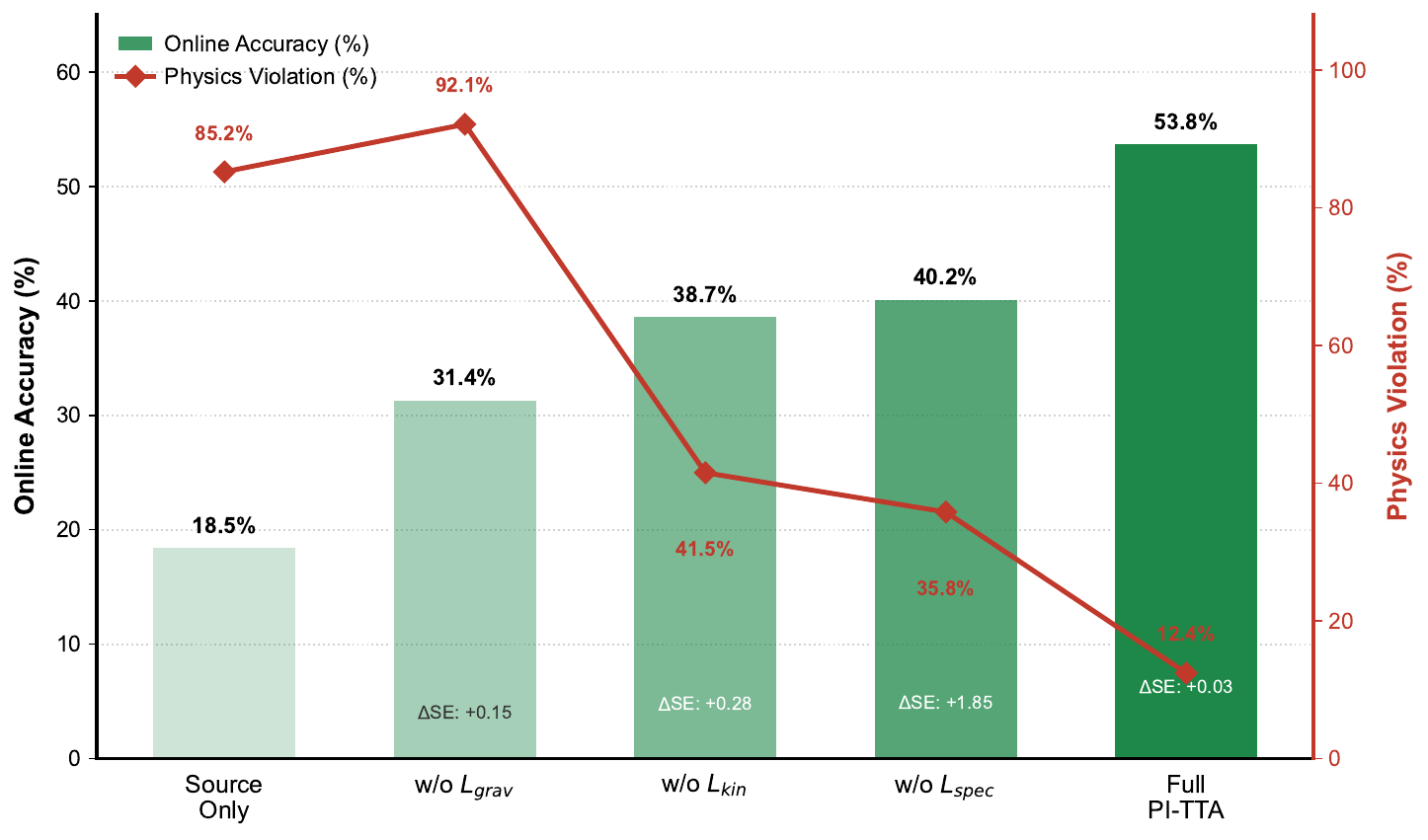}%
  \label{fig:ablation_a}}%
\hfill
\subfloat[Sensitivity to constraint weights $\lambda$]{%
  \includegraphics[width=0.82\linewidth]{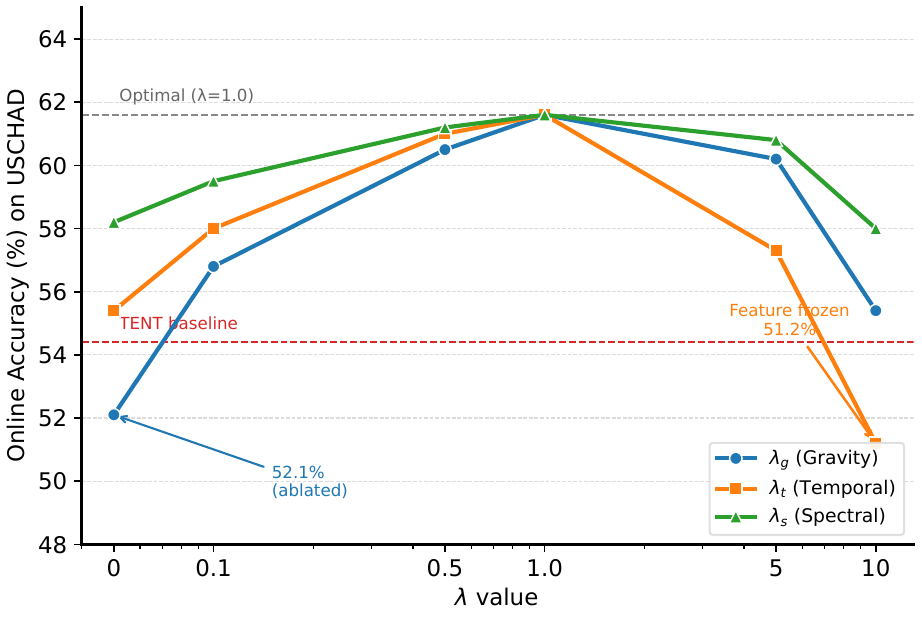}%
  \label{fig:ablation_b}}%
\caption{Ablation and sensitivity analysis of \textsc{PI-TTA}.}
\label{fig:ablation_overhead}
\end{figure}

We next test whether \textsc{PI-TTA} behaves according to its intended design and whether the observed gains are consistent with the proposed mechanism rather than a generic regularization effect. Fig.~\ref{fig:ablation_overhead} reports the ablation and sensitivity results.

The ablation results confirm the intended division of labor among the three constraints. Gravity consistency contributes most strongly under rotation shifts, short-horizon temporal continuity suppresses instability in long correlated streams, and spectral stability is especially important under sampling-rate drift. This alignment supports the interpretation that each component addresses a distinct degradation mode rather than acting as a generic regularizer. In Fig.~\ref{fig:ablation_a}, SE denotes the end-stage spectral entropy diagnostic.

We further examine the gravity proxy itself, since batch-level mean acceleration may be contaminated by body acceleration during highly dynamic motion. In quasi-static activities, the gravity term remains highly effective. Under highly dynamic activities, its precision decreases because stronger body acceleration perturbs the proxy. The proposed variance-based gate mitigates this issue by reducing the effective gravity weight in unreliable batches and preserving much of the benefit.

\subsection{Failure Visualization: The Low-Entropy Trap Versus Physical Feasibility}

\begin{figure}[t]
\centering
\subfloat[Feature-space evolution]{%
  \includegraphics[width=0.85\linewidth]{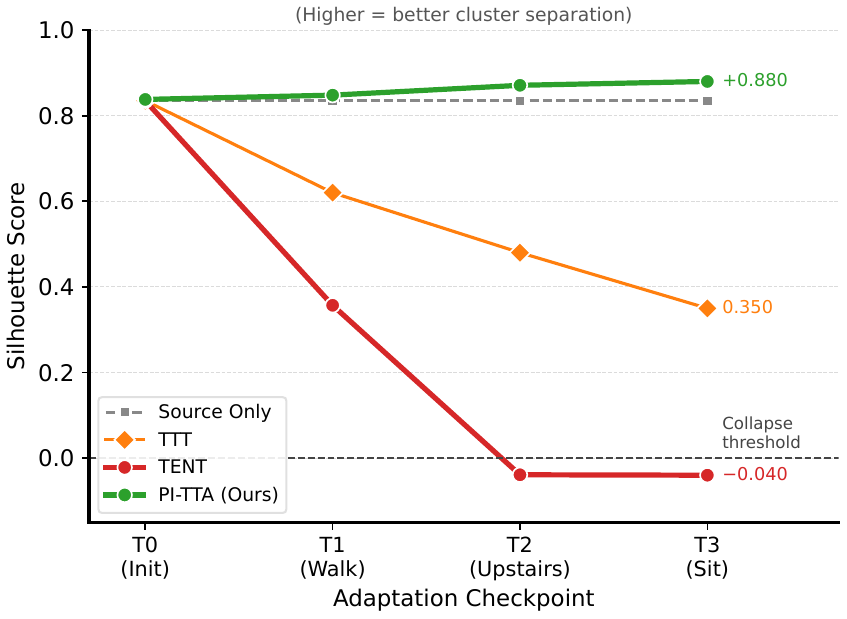}%
  \label{fig:silhouette}} \\[4pt]

\subfloat[Gravity-feasibility ribbon]{%
  \includegraphics[width=0.85\linewidth]{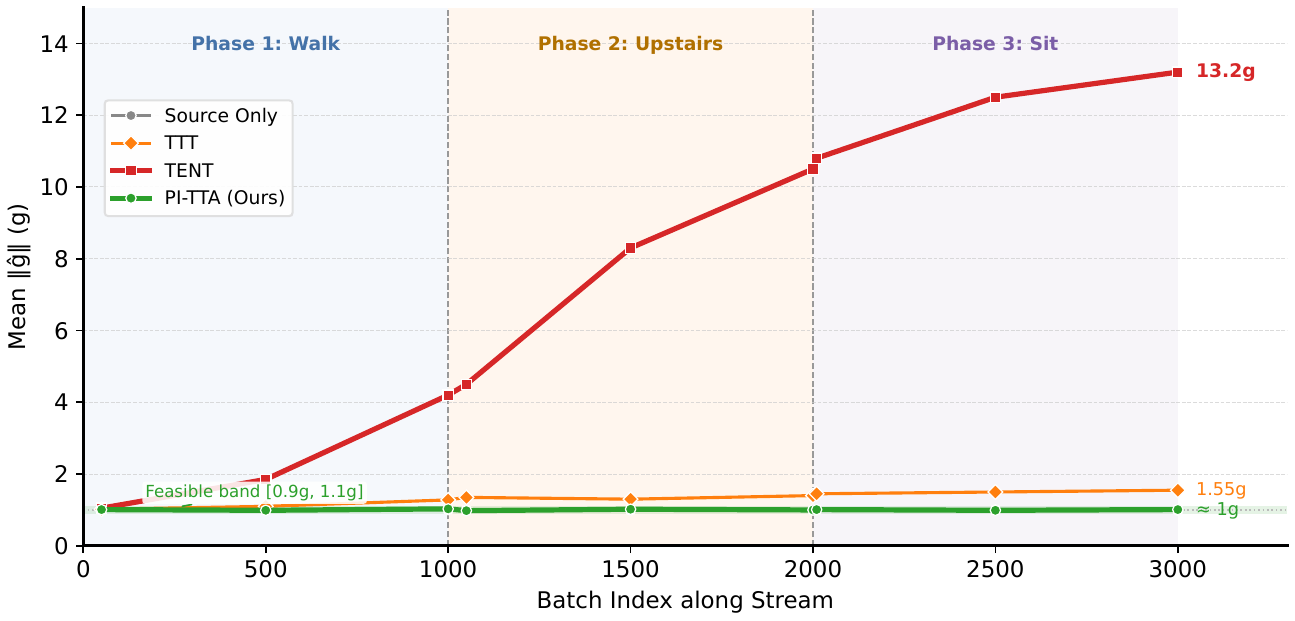}%
  \label{fig:ribbon}} \\[4pt]

\subfloat[Entropy versus physical-violation]{%
  \includegraphics[width=0.85\linewidth]{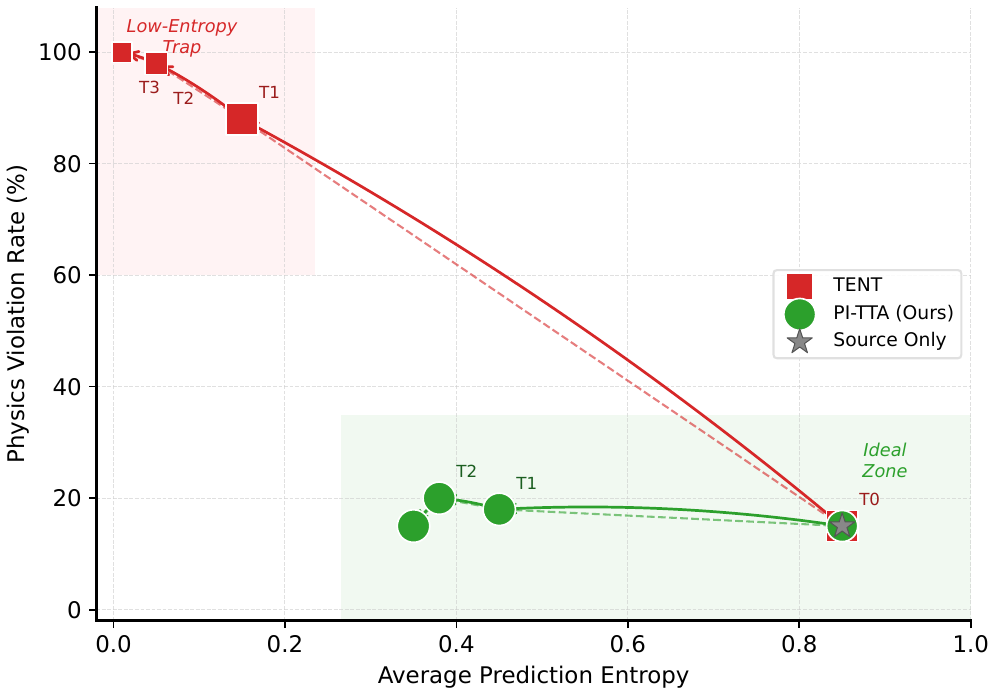}%
  \label{fig:entropy_viol}}

\caption{Failure visualization under streaming adaptation.}
\label{fig:failure_viz}
\end{figure}

We next provide a visual validation of the failure mechanisms discussed earlier. Scalar accuracy alone is insufficient to explain why streaming adaptation succeeds or fails under temporally correlated sensor streams. We therefore examine three complementary diagnostics: a gravity-feasibility ribbon, feature-space evolution with t-SNE checkpoints and silhouette score, and the relation between prediction entropy and physical-violation rate~\cite{gong2022note,gong2023sotta,zhao2023ttab}. Fig.~\ref{fig:failure_viz} summarizes these diagnostics.

\textbf{Gravity-feasibility ribbon.}
As shown in Fig.~\ref{fig:ribbon}, entropy reduction under \textsc{TENT} can be achieved by drifting into physically implausible states rather than by recovering a valid sensor representation. We observe broad deviations, together with extreme outliers in which the gravity-proxy magnitude rises to 13.2\,g on USCHAD and 12.39\,g on mHealth. By contrast, \textsc{PI-TTA} maintains a tight feasibility ribbon centered around 1.0\,g throughout the stream.

\textbf{Feature-space evolution.}
Fig.~\ref{fig:silhouette} shows that under \textsc{TENT}, the representation progresses from initially separated clusters to strong inter-class mixing and final collapse, with the silhouette score dropping from $0.835$ at $T_0$ to $-0.040$ at $T_3$ on USCHAD. In contrast, \textsc{PI-TTA} preserves the global class geometry and improves compactness over time, reaching $\mathbf{0.880}$ at $T_3$.

\textbf{Entropy versus physical-violation.}
The trajectories in Fig.~\ref{fig:entropy_viol} reveal a characteristic scissor pattern under \textsc{TENT}: prediction entropy decreases monotonically, yet physical-violation increases sharply. Lower entropy alone is therefore not a reliable indicator of successful adaptation on correlated inertial streams. By contrast, \textsc{PI-TTA} avoids this failure mode by maintaining lower violation with stable accuracy.

To quantify physical plausibility, we define the \emph{physical-violation rate} as
\begin{equation}
\mathrm{VR} \;=\; \frac{1}{T}\sum_{t=1}^{T}\mathbb{I}\!\left[\|\hat{\mathbf g}_t\| \notin [0.9\,\mathrm{g},1.1\,\mathrm{g}]\right],
\end{equation}
where $\|\hat{\mathbf g}_t\|$ denotes the gravity-proxy magnitude at time $t$. Relative to entropy minimization, \textsc{PI-TTA} reduces the violation rate by \textbf{27.5\%} on USCHAD, \textbf{24.1\%} on PAMAP2, and \textbf{45.4\%} on mHealth. These results show that the benefit of \textsc{PI-TTA} is not limited to scalar accuracy. It also reduces the tendency to exchange physical plausibility for spurious confidence.

\subsection{Fairness, Robustness, and Generalization Analysis}

We next test whether the observed gains are tied to a specific backbone or a narrowly tuned operating point. To this end, we replace the default encoder with architecturally distinct alternatives, including DeepConvLSTM and a lightweight transformer backbone, while keeping the long-sequence streaming protocol, the adaptation interface, and the optimization settings unchanged. Table~\ref{tab:backbone_generalization} summarizes this backbone comparison. All models are evaluated under the same class-sorted long-sequence setting, and lower peak gravity-proxy magnitude indicates better physical plausibility.

\begin{table}[t]
\centering
\caption{Backbone generalization under the long-sequence protocol.}
\label{tab:backbone_generalization}
\fontsize{7}{8}\selectfont
\setlength{\tabcolsep}{4pt}
\resizebox{\linewidth}{!}{
\begin{tabular}{lccccc}
\toprule
\textbf{Backbone} & \textbf{Params} & \textbf{Source Only} & \textbf{\textsc{TENT}} & \textbf{\textsc{PI-TTA}} & \textbf{Peak gravity-proxy magnitude $\|\hat{\mathbf g}_t\|$} \\
\midrule
DeepConvLSTM & 1.2M & 50.4 $\pm$ 1.2 & 10.8 $\pm$ 1.5 & \textbf{59.8 $\pm$ 0.9} & 1.04\,g \\
1D-ResNet18 (default) & 3.8M & 51.6 $\pm$ 0.8 & 12.4 $\pm$ 1.5 & \textbf{61.6 $\pm$ 0.7} & 1.03\,g \\
Mobile-Transformer & 2.5M & 53.2 $\pm$ 1.0 & 18.5 $\pm$ 2.2 & \textbf{60.9 $\pm$ 0.8} & 1.06\,g \\
\bottomrule
\end{tabular}}
\end{table}

Table~\ref{tab:backbone_generalization} shows a consistent pattern across all three architectures. Confidence-driven adaptation remains unstable, whereas \textsc{PI-TTA} consistently preserves strong terminal accuracy and keeps the gravity-proxy magnitude close to the feasible regime.

We finally test whether the observed gains could be explained primarily by hyperparameter favoritism rather than by the method itself. Table~\ref{tab:fairness_lr} reports terminal long-stream accuracy under a shared learning-rate grid, with values measured as Phase-3 accuracy under the long-sequence protocol.

\begin{table}[t]
\centering
\caption{Fairness analysis under a shared learning-rate grid.}
\label{tab:fairness_lr}
\fontsize{7}{8}\selectfont
\setlength{\tabcolsep}{5pt}
\begin{tabular}{lcccc}
\toprule
\textbf{Learning Rate} & \textbf{\textsc{TENT}} & \textbf{\textsc{EATA}} & \textbf{\textsc{CoTTA}} & \textbf{\textsc{PI-TTA}} \\
\midrule
$1\times 10^{-4}$ & 45.2 & 48.5 & 46.2 & \textbf{58.5} \\
$1\times 10^{-3}$ & 12.4 & 28.4 & 45.3 & \textbf{61.6} \\
$1\times 10^{-2}$ & 5.2 & 15.6 & 42.1 & \textbf{59.2} \\
\bottomrule
\end{tabular}
\end{table}

Entropy-driven methods are highly sensitive to tuning, with \textsc{TENT} collapsing as the learning rate increases and \textsc{EATA} degrading substantially once updates become more aggressive. \textsc{CoTTA} is more stable, but its accuracy remains consistently below that of \textsc{PI-TTA}. By contrast, \textsc{PI-TTA} remains near the 60\% regime across the entire grid, indicating stronger out-of-the-box robustness under practical tuning uncertainty.

Overall, these results show that the gains of \textsc{PI-TTA} remain robust beyond a single backbone or a narrowly tuned operating point.

\subsection{Deployment Trade-offs and Sustained On-Device Stability Under Resource Budgets}

We next profile the deployment behavior of \textsc{PI-TTA} under constrained update opportunities and limited on-device compute budgets. This setting is practically important because test-time adaptation on mobile devices must compete with foreground workloads, sensing pipelines, thermal limits, and power constraints~\cite{alfarra2024ttatime,shi2016edge,lim2020fedcomst}. We therefore examine five complementary aspects: single-step profiling, accuracy--latency trade-offs, effective accuracy under strict time budgets, absolute energy--latency behavior under sparse updates, and sustained on-device stability during continuous execution.

We begin with single-step profiling. Table~\ref{tab:overhead} reports on-device overhead on Snapdragon~8~Gen~2 with batch size $=64$ and a 1D-ResNet backbone. All values are measured as mean $\pm$ standard deviation over 5 independent runs, and the updated-parameter ratio is reported relative to the total backbone parameters. Lower values in latency, memory, and updated-parameter ratio indicate better deployability.

\begin{table}[t]
\centering
\caption{On-device overhead profiling on Snapdragon~8~Gen~2.}
\label{tab:overhead}
\fontsize{7}{8}\selectfont
\resizebox{1.0\linewidth}{!}{
\begin{tabular}{lcccc}
\toprule
\textbf{Method} & 
\makecell{\textbf{Latency / step} \\ \textbf{(ms)}} & 
\makecell{\textbf{Peak memory} \\ \textbf{(MB)}} & 
\makecell{\textbf{Updated params} \\ \textbf{(\%)}} & 
\makecell{\textbf{T3 Accuracy} \\ \textbf{(\%)}} \\
\midrule
Source Only & \textbf{15.2 $\pm$ 0.4} & \textbf{18.5 $\pm$ 0.2} & \textbf{0} & 51.6 $\pm$ 0.8 \\
TENT~\cite{wang2021tent} & 38.5 $\pm$ 1.1 & \textbf{26.0 $\pm$ 0.5} & \textbf{${\sim}2.5$} & 12.4 $\pm$ 1.5 \\
TTT~\cite{sun2020ttt} & 82.4 $\pm$ 2.3 & 58.2 $\pm$ 1.8 & ${\sim}5.4$ & 17.1 $\pm$ 2.1 \\
\textsc{PI-TTA} (Ours) & 45.1 $\pm$ 1.2 & 30.4 $\pm$ 0.6 & ${\sim}2.9$ & \textbf{61.6 $\pm$ 0.7} \\
\bottomrule
\end{tabular}}
\end{table}

Table~\ref{tab:overhead} shows that \textsc{PI-TTA} does incur additional cost relative to inference-only execution, as expected for any adaptive method, but the overhead remains moderate relative to stronger self-supervised baselines. In particular, \textsc{PI-TTA} is only slightly heavier than \textsc{TENT} in latency and memory, while remaining far more efficient than \textsc{TTT}. At the same time, it substantially outperforms both adaptive baselines in terminal long-stream accuracy.

\begin{figure}[t]
\centering
\subfloat[Accuracy--latency Pareto frontier]{%
  \includegraphics[width=0.76\linewidth]{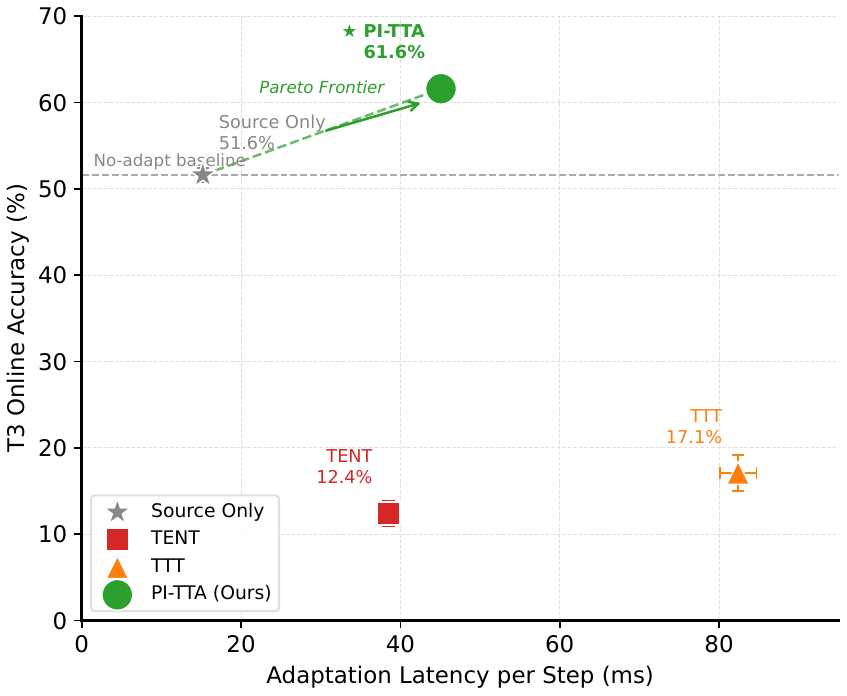}%
  \label{fig:pareto_a}}
\vspace{0.3cm}
\subfloat[Update interval $K$ versus accuracy and energy]{%
  \includegraphics[width=0.78\linewidth]{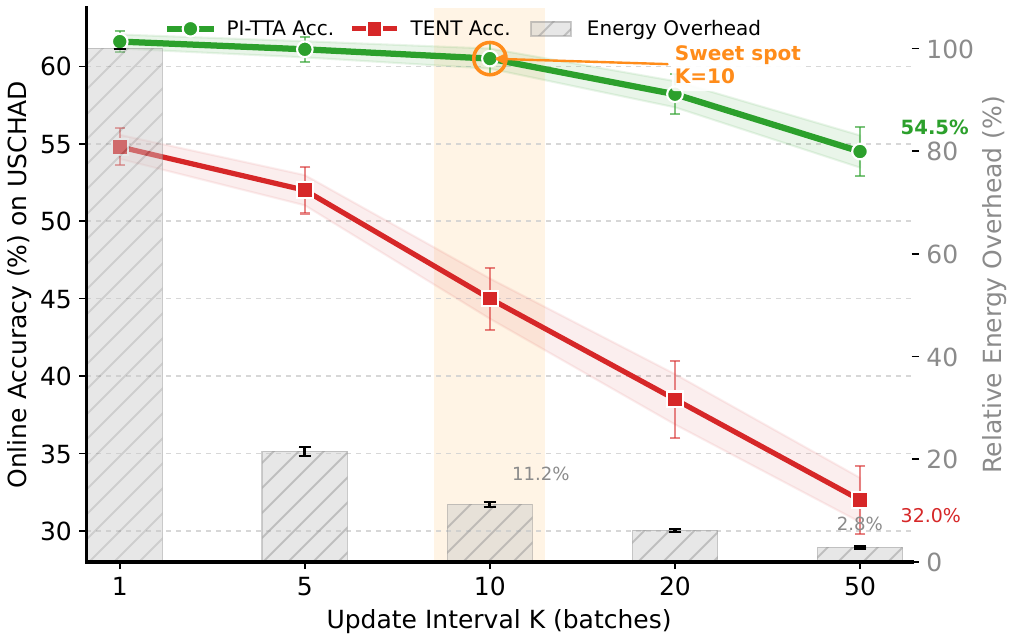}%
  \label{fig:pareto_b}}
\caption{Deployment trade-offs under constrained update budgets.}
\label{fig:pareto}
\end{figure}

Fig.~\ref{fig:pareto} summarizes the deployment trade-offs from two complementary perspectives. Fig.~\ref{fig:pareto_a} reports the accuracy--latency Pareto frontier across adaptation methods, while Fig.~\ref{fig:pareto_b} studies how the update interval $K$ controls the energy--accuracy balance under sparse update schedules.

The Pareto frontier in Fig.~\ref{fig:pareto_a} shows that \textsc{PI-TTA} achieves substantially stronger adaptation quality than confidence-driven baselines while remaining within a practical latency range for on-device execution. The update-interval study in Fig.~\ref{fig:pareto_b} further clarifies the deployment boundary. As $K$ increases, all adaptive methods reduce energy consumption because backward updates are triggered less frequently. However, the performance of entropy-driven adaptation deteriorates much more sharply, whereas \textsc{PI-TTA} degrades gracefully and preserves strong accuracy even under sparse update schedules. In our setting, $K{=}10$ emerges as a practical operating point because it is the smallest update interval that satisfies the strict 20\,ms budget while preserving near-peak accuracy.

\paragraph{Time-Constrained Measurement Protocol}
Standard TTA evaluation assumes an idealized synchronous pipeline in which adaptation completes before the next batch arrives. In mobile sensing, however, the available adaptation time is bounded by the inter-arrival interval of streaming windows. For example, a 20\,Hz stream implies a 50\,ms budget, whereas a 50\,Hz stream implies a strict 20\,ms budget. To capture real deployment viability, we therefore enforce a \emph{time-constrained scoring protocol}. Let $T_{\text{adapt}}$ denote the measured adaptation latency and let $T_{\text{budget}}$ denote the inter-arrival budget. If $T_{\text{adapt}} \leq T_{\text{budget}}$, the update is applied before the next window and is marked \emph{safe}. If $T_{\text{budget}} < T_{\text{adapt}} \leq 2T_{\text{budget}}$, the update is \emph{delayed by one cycle}. If $T_{\text{adapt}} > 2T_{\text{budget}}$, one or more adaptation opportunities are missed and the update is marked \emph{dropped}. Effective accuracy is measured under the resulting delayed-update schedule rather than under an idealized synchronous assumption.

Table~\ref{tab:time_constrained} reports the resulting effective accuracy under strict time budgets. Heavier methods such as TTT and CoTTA exceed the budget by a large margin and therefore suffer a clear drop from their idealized synchronous accuracy once delay and dropped-update effects are taken into account. Even dense \textsc{PI-TTA} updates with $K{=}1$ begin to degrade under the strict 20\,ms budget. By contrast, \textsc{PI-TTA} with a sparse schedule ($K{=}10$) averages 18.1\,ms per step, fits within the 50\,Hz deadline, and maintains its 60.5\% effective accuracy without real-time penalty. This identifies a practically relevant operating point in which robust adaptation coexists with strict mobile sensing deadlines.

\begin{table}[t]
\centering
\caption{Effective accuracy under time-constrained streaming.}
\label{tab:time_constrained}
\fontsize{7}{8}\selectfont
\setlength{\tabcolsep}{4pt}
\resizebox{\linewidth}{!}{
\begin{tabular}{lccc}
\toprule
\multirow{2}{*}{\textbf{Method}} & \textbf{Idealized Sync} & \textbf{Budget: 50\,ms} & \textbf{Budget: 20\,ms} \\
& \textbf{Acc (\%)} & \textbf{(20\,Hz Sensor)} & \textbf{(50\,Hz Sensor)} \\
\midrule
Source Only & 51.5 & 51.5 & 51.5 \\
\midrule
TTT~\cite{sun2020ttt} & 17.1 & 15.2 (delayed by one cycle) & 12.8 (dropped) \\
CoTTA~\cite{Wang_2022_CVPR_CoTTA} & 45.3 & 42.1 (delayed by one cycle) & 38.5 (dropped) \\
\midrule
\textsc{PI-TTA} ($K=1$) & \textbf{61.6} & \textbf{61.6} (safe) & 58.2 (delayed by one cycle) \\
\textsc{PI-TTA} ($K=10$) & 60.5 & 60.5 (safe) & \textbf{60.5} (safe) \\
\bottomrule
\end{tabular}}
\end{table}

To complement the relative energy view in Fig.~\ref{fig:pareto_b}, Table~\ref{tab:absolute_energy_latency} reports absolute per-batch energy and latency under different update intervals on Snapdragon~8~Gen~2. Lower values indicate lower energy cost and shorter runtime. At $K{=}10$, \textsc{PI-TTA} reduces the average latency to 18.1\,ms, placing it within the 20\,ms timing budget commonly associated with 50\,Hz real-time mobile sensing, while preserving 60.5\% terminal accuracy. By contrast, very sparse updates such as $K{=}50$ further reduce energy and latency, but begin to incur a clearer accuracy penalty.

\begin{table}[t]
\centering
\caption{Absolute energy and latency under different update intervals.}
\label{tab:absolute_energy_latency}
\fontsize{7}{8}\selectfont
\setlength{\tabcolsep}{4pt}
\resizebox{\linewidth}{!}{
\begin{tabular}{lcccc}
\toprule
\textbf{Update} & \textbf{Absolute Energy} & \textbf{Avg.\ Latency} & \textbf{P99 Latency} & \textbf{Terminal Acc.} \\
\textbf{Interval $K$} & \textbf{(mJ/batch)} & \textbf{(ms)} & \textbf{(ms)} & \textbf{(T3)} \\
\midrule
1  & 28.2 $\pm$ 1.0 & 45.1 $\pm$ 1.2 & 55.8 & 61.6 $\pm$ 0.7 \\
5  & 13.5 $\pm$ 0.6 & 21.2 $\pm$ 0.8 & 25.4 & 61.1 $\pm$ 0.8 \\
10 & \textbf{9.2 $\pm$ 0.4} & \textbf{18.1 $\pm$ 0.5} & \textbf{20.2} & 60.5 $\pm$ 1.0 \\
20 & 6.8 $\pm$ 0.3 & 16.7 $\pm$ 0.4 & 18.5 & 58.7 $\pm$ 1.1 \\
50 & 4.8 $\pm$ 0.2 & 15.8 $\pm$ 0.3 & 17.2 & 55.4 $\pm$ 1.3 \\
\bottomrule
\end{tabular}}
\end{table}

Static profiling alone is insufficient for mobile deployment. A method may appear efficient over a short benchmark window yet become unstable during sustained execution because of thermal throttling, memory growth, or tail-latency spikes. We therefore perform a 30-minute on-device sustained execution test on Snapdragon~8~Gen~2 under continuous single-threaded streaming inference and adaptation. Table~\ref{tab:stress_test} summarizes this sustained-run behavior. Lower values indicate better sustained efficiency and thermal stability.

\begin{table}[t]
\centering
\caption{Sustained on-device stability under 30-minute continuous execution.}
\label{tab:stress_test}
\fontsize{7}{8}\selectfont
\setlength{\tabcolsep}{4pt}
\resizebox{\linewidth}{!}{
\begin{tabular}{lccccc}
\toprule
\textbf{Method} & \makecell{\textbf{Run} \\ \textbf{Time}} & 
\makecell{\textbf{Avg. Latency} \\ \textbf{(ms)}} & 
\makecell{\textbf{P99 Latency} \\ \textbf{(ms)}} & 
\makecell{\textbf{Peak Memory} \\ \textbf{(MB)}} & 
\makecell{\textbf{Chip Temp.} \\ \textbf{($^\circ$C)}} \\
\midrule
TTT~\cite{sun2020ttt} & 5 min  & 82.4  & 95.2  & 58.2  & 36.5 \\
TTT~\cite{sun2020ttt} & 15 min & 115.6 & 142.5 & 58.5  & 42.1 \\
TTT~\cite{sun2020ttt} & 30 min & 185.2 & 250.4 & 59.0  & 46.8 \\
\midrule
CoTTA~\cite{Wang_2022_CVPR_CoTTA} & 5 min  & 65.5 & 185.2 & 75.4  & 35.8 \\
CoTTA~\cite{Wang_2022_CVPR_CoTTA} & 30 min & 82.1 & 350.5 & 110.5 & 39.5 \\
\midrule
\textsc{PI-TTA} & 5 min  & \textbf{45.1} & \textbf{52.3} & \textbf{30.4} & 33.2 \\
\textsc{PI-TTA} & 15 min & \textbf{45.5} & \textbf{54.1} & \textbf{30.4} & 35.5 \\
\textsc{PI-TTA} & 30 min & \textbf{46.2} & \textbf{55.8} & \textbf{30.5} & \textbf{38.2} \\
\bottomrule
\end{tabular}}
\end{table}

Table~\ref{tab:stress_test} reveals a clear difference between short-run efficiency and sustained usability. TTT becomes progressively slower as execution continues, with average latency rising from 82.4\,ms at 5 minutes to 185.2\,ms at 30 minutes, accompanied by a large increase in P99 latency and chip temperature. CoTTA is more moderate in average latency, but it exhibits much larger tail latency and substantial memory growth over time. By contrast, \textsc{PI-TTA} remains stable across the full 30-minute run. Its average latency changes only marginally, P99 latency stays tightly bounded, peak memory remains essentially constant, and chip temperature increases only moderately.

These results show that \textsc{PI-TTA} remains useful not only under standard evaluation, but also under the latency, energy, thermal, and real-time deadline constraints that matter in sustained mobile deployment.

\subsection{Limitations and Failure Cases}

\textsc{PI-TTA} is not intended as a universally valid adaptation rule. Its effectiveness depends on the continued usefulness of lightweight physical proxies, including gravity consistency, short-horizon temporal continuity, and spectral stability. When sensing conditions strongly violate these assumptions, such as under severe sampling-rate distortion, sensor saturation, prolonged high-dynamic motion bursts, or acquisition irregularities that distort the inertial structure itself~\cite{kwon2020imutube,haresamudram2020maskedhar,zhao2023ttab,alfarra2024ttatime}, the corresponding constraints become less reliable and adaptation quality can degrade.

This boundary is already visible in the factorized shift analysis. Under extreme drift, \textsc{TTT} can approach \textsc{PI-TTA} in some regimes, suggesting that geometric self-supervision and spectral stabilization may provide partially complementary signals near the boundary of severe temporal distortion. Likewise, the gravity proxy is intentionally down-weighted under highly dynamic motion because batch-level mean acceleration is not a precise gravity estimator in that regime. These cases do not invalidate the method; rather, they clarify that \textsc{PI-TTA} is most effective when lightweight physical cues remain informative enough to stabilize online updates. At the same time, \textsc{PI-TTA} focuses on physics-consistent stabilization rather than explicit reset, replay, or buffering strategies; integrating such mechanisms may further improve recovery under extreme non-stationary bursts.

In practice, such regimes can be handled by reliability-aware fallback strategies, including constraint down-weighting, temporary adaptation pause, or inference-only fallback until the stream returns to a reliable operating condition. An important direction for future work is therefore reliability-aware adaptation control, in which the method decides not only how to adapt, but also when adaptation should be weakened or suspended.
\section{Conclusion}
We presented \textsc{PI-TTA}, a lightweight physics-informed framework for source-free test-time adaptation on behavioral inertial streams for mobile human activity recognition. By augmenting standard lightweight adaptation with gravity consistency, short-horizon temporal continuity, and spectral stability, \textsc{PI-TTA} stabilizes online updates under temporally correlated mobile deployment conditions. Experiments on USCHAD, PAMAP2, and mHealth show that \textsc{PI-TTA} improves long-sequence adaptation stability, preserves physically plausible behavior under realistic streaming shifts, and remains compatible with resource-constrained on-device execution. These results demonstrate that lightweight physical structure can serve as an effective inductive bias for deployable source-free adaptation in mobile sensing systems.

More broadly, our findings suggest that deployable adaptation for behavioral streams should be treated as a systems problem that jointly considers adaptation reliability, temporal stability, and runtime efficiency. \textsc{PI-TTA} provides a practical step toward robust source-free on-device adaptation for real-world wearable and mobile HAR.

\end{document}